\documentclass[11pt,a4paper]{article}
\usepackage[hyperref]{acl2021}
\usepackage{times}
\usepackage{latexsym}

\usepackage[disable]{todonotes}
\usepackage{microtype}
\usepackage{booktabs}
\usepackage{mathrsfs}
\usepackage{makecell}
\usepackage{multirow}
\usepackage{graphicx}

\usepackage{subcaption}
\usepackage{graphicx}
\usepackage{subcaption}
\usepackage{mwe}

\usepackage{booktabs}
\usepackage{mathrsfs}
\usepackage{makecell}
\usepackage{multirow}
\usepackage{graphicx}
\usepackage{subcaption}
\usepackage{amsthm}
\usepackage{amsmath}
\usepackage{amsfonts}
\usepackage{bm}
\usepackage{soul}
\usepackage{subcaption,siunitx,booktabs}
\newcommand{\metric}[1]{\textsc{#1}}
\usepackage{amsmath,amsfonts,amssymb,amsthm}

\newcommand{\insertDistLanguages}{
\begin{figure*}[h!]
\begin{centering}
\includegraphics[width=0.67\linewidth]{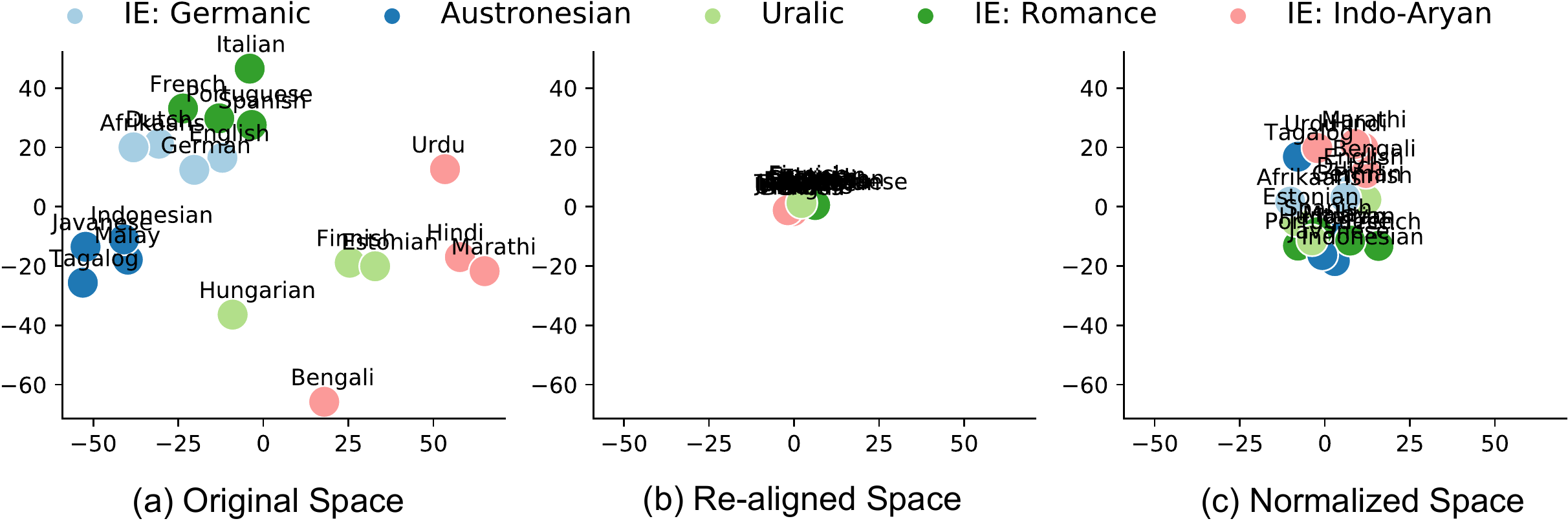}
 \caption{
 t-SNE distributions of language centroids based on the last m-BERT layer.
 }
 \label{fig:vis_lanaguges}  
\end{centering}
\end{figure*} 
}

\newcommand{\insertPerfGains}{
\begin{figure}
\centerline{\includegraphics[width=0.8\linewidth]{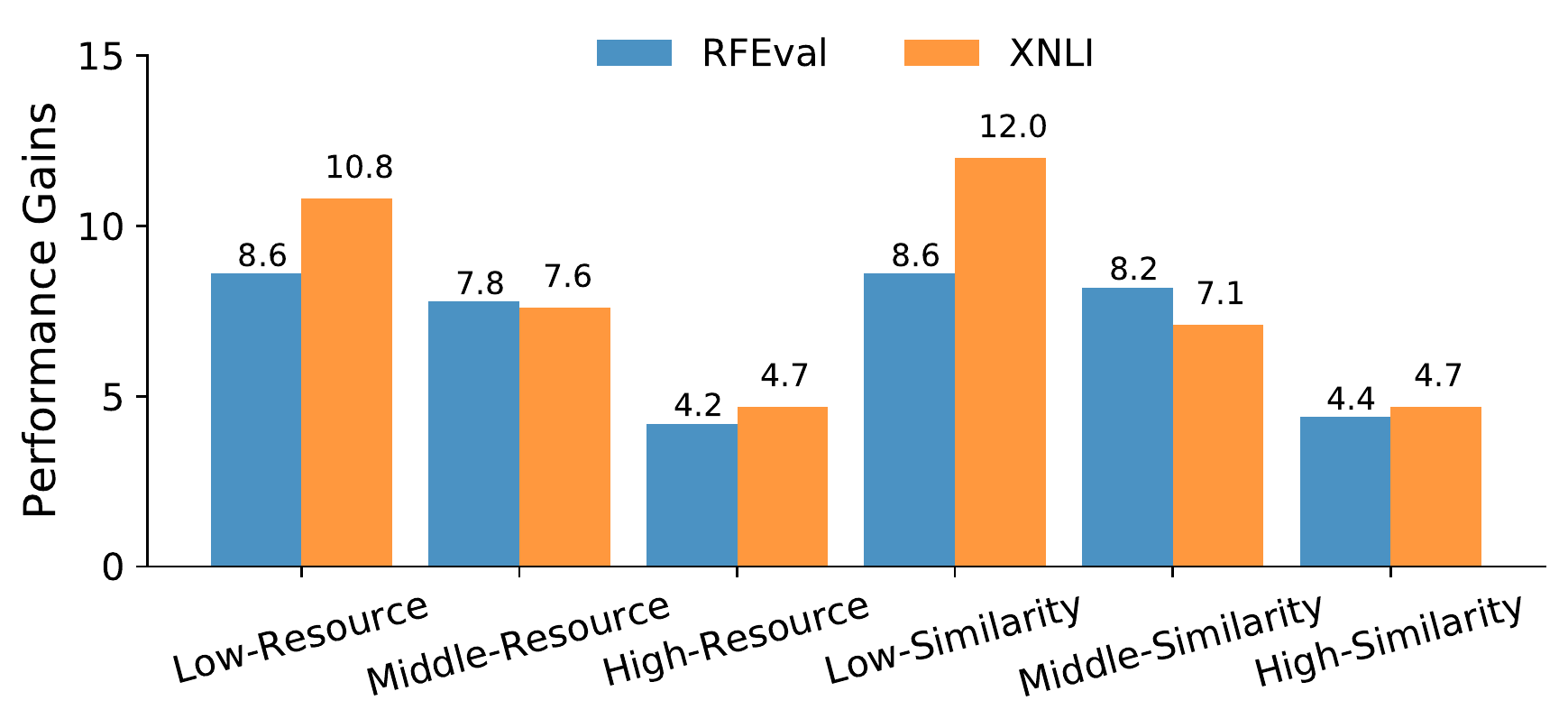}}  
 \caption{Performance gains across language groups 
 for \metric{m-BERT $\oplus$ Joint-Align $\oplus$ Norm}.
 }
 \label{fig:perf_gains}  
\end{figure} 
}

\newcommand{\insertPerfHuman}{
\begin{figure}
\centerline{\includegraphics[width=0.75\linewidth]{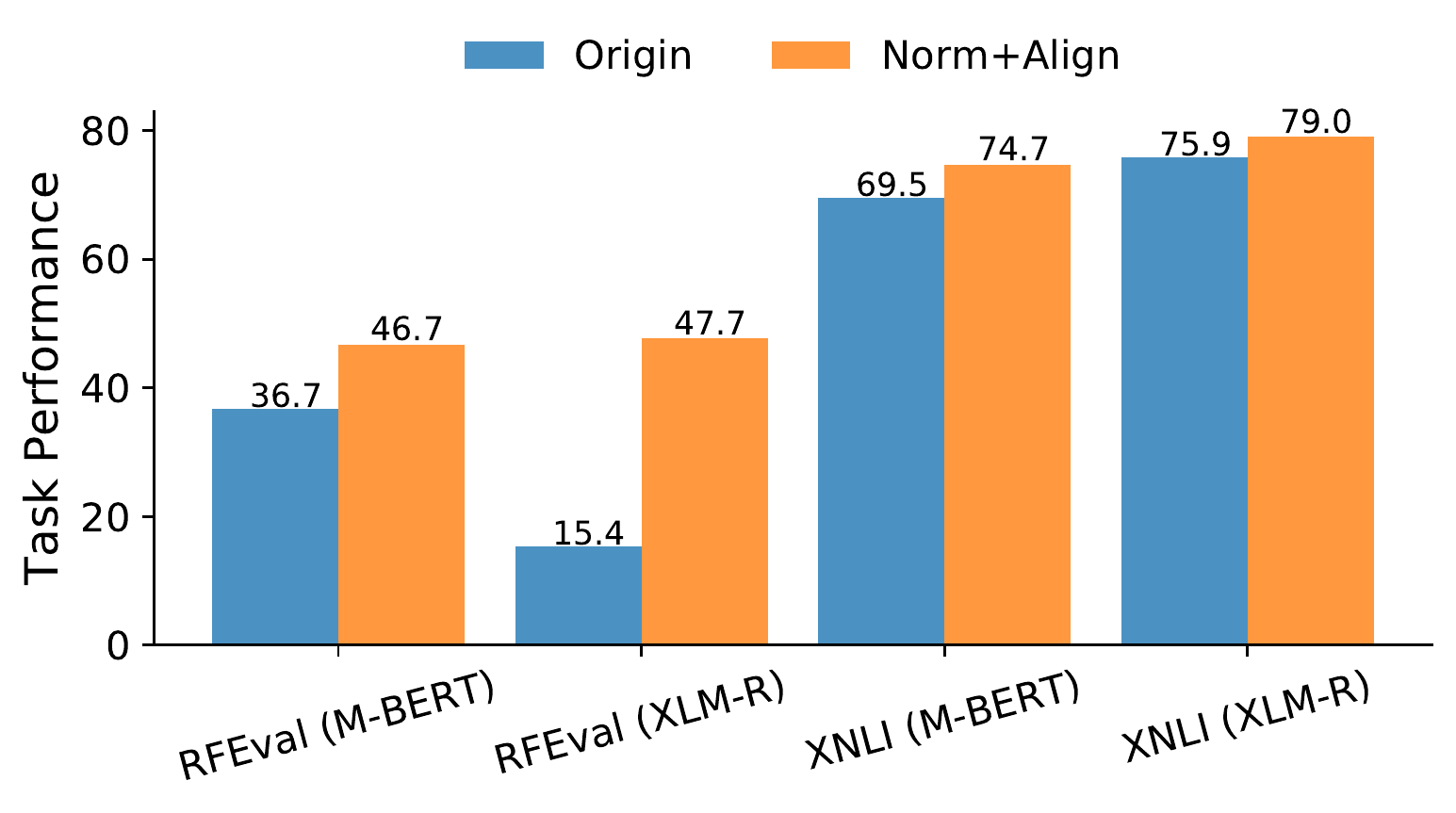}}  
 \caption{Results on \rf{} are averaged over two language pairs (de-en and fi-en) from the WMT17 human translated test sets. Likewise, results on \xnli{} are averaged over four selected language pairs (en-fr, en-de, en-hi and en-es) from \xnli{} human translated test sets.
 }
 \label{fig:perf_human}  
\end{figure} 
}

\newcommand{\insertDistplots}{
\begin{figure*}
\begin{minipage}{0.33\textwidth}  
	\centerline{\includegraphics[width=0.85\linewidth]{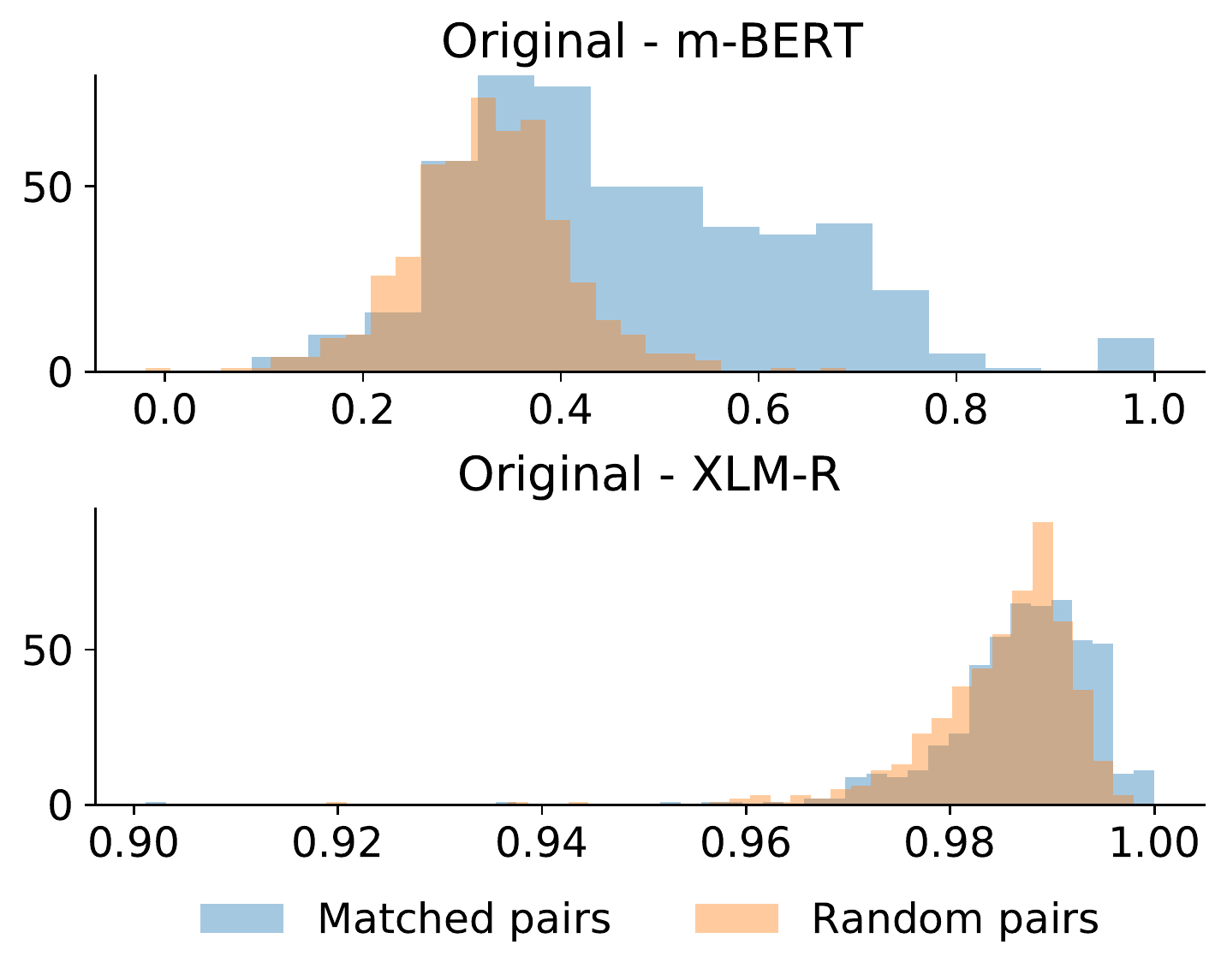}}  
\end{minipage} 
\begin{minipage}{0.33\textwidth}  
	\centerline{\includegraphics[width=0.85\linewidth]{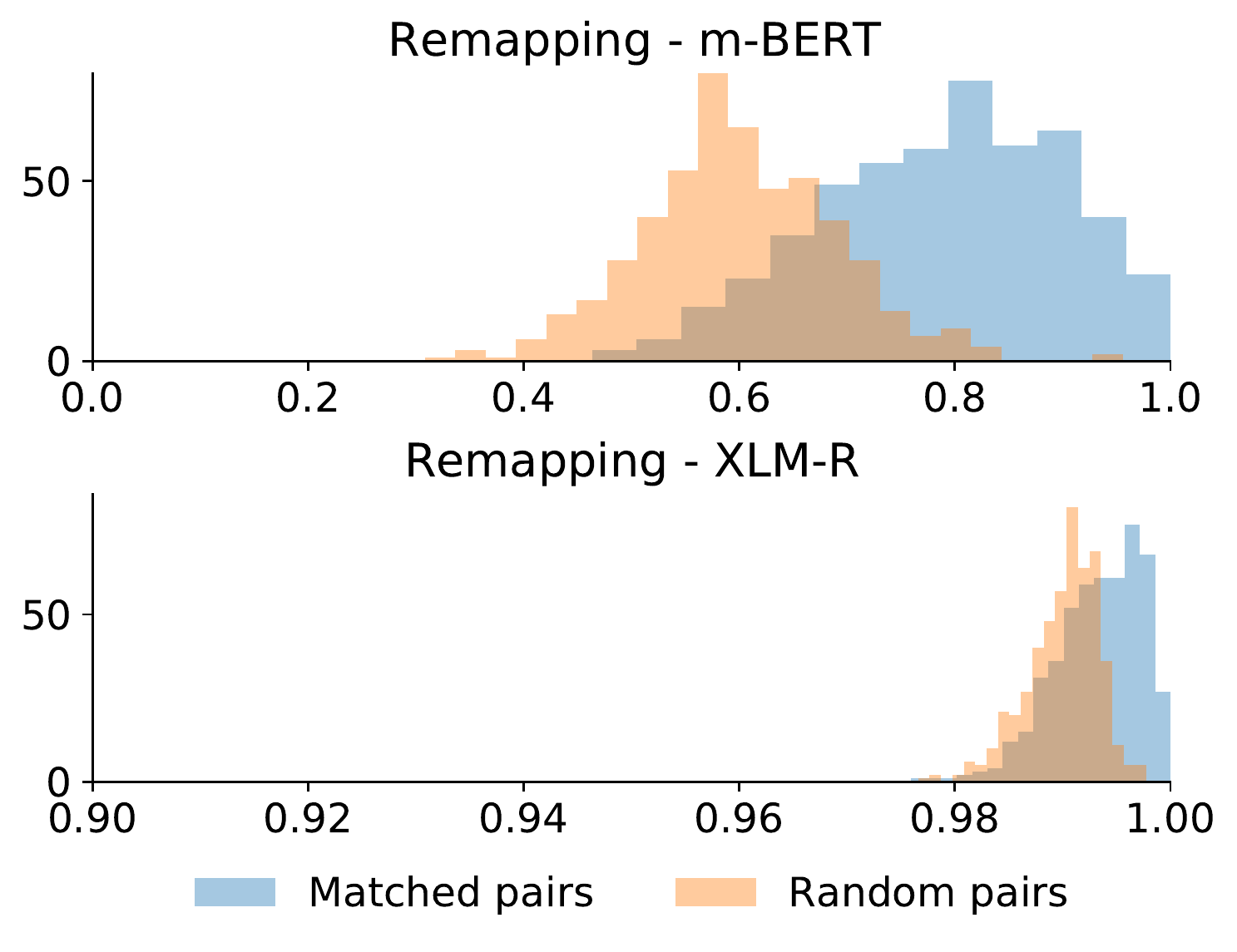}}  
\end{minipage} 
\begin{minipage}{0.33\textwidth}  
	\centerline{\includegraphics[width=0.85\linewidth]{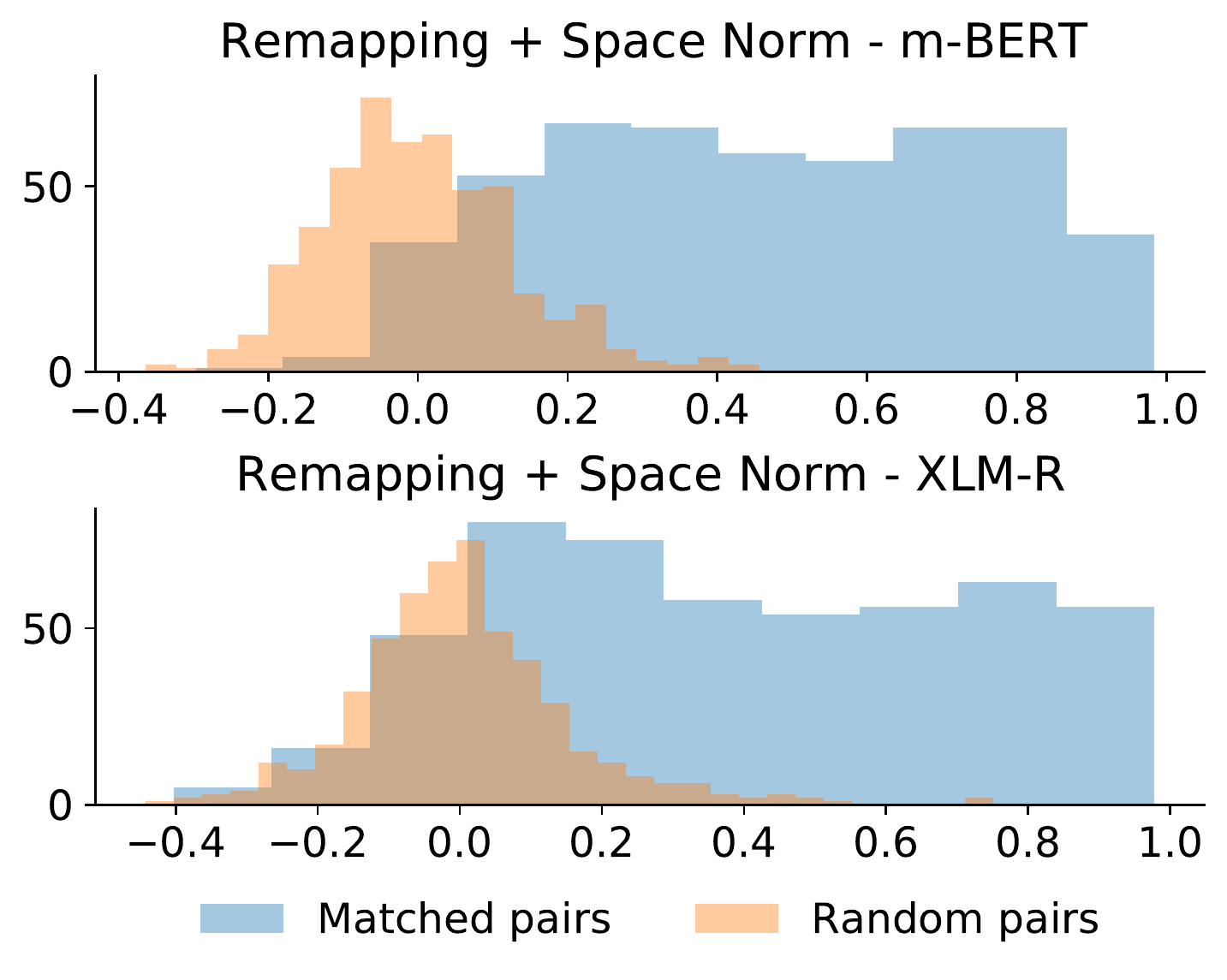}}
\end{minipage} 
\caption{Histograms of cosine similarity scores of word pairs.
}
\label{fig:histogram}
\end{figure*}  
}

\newcommand{\insertBatchNorm}{
\begin{figure}
\centering
\includegraphics[width=0.8\linewidth]{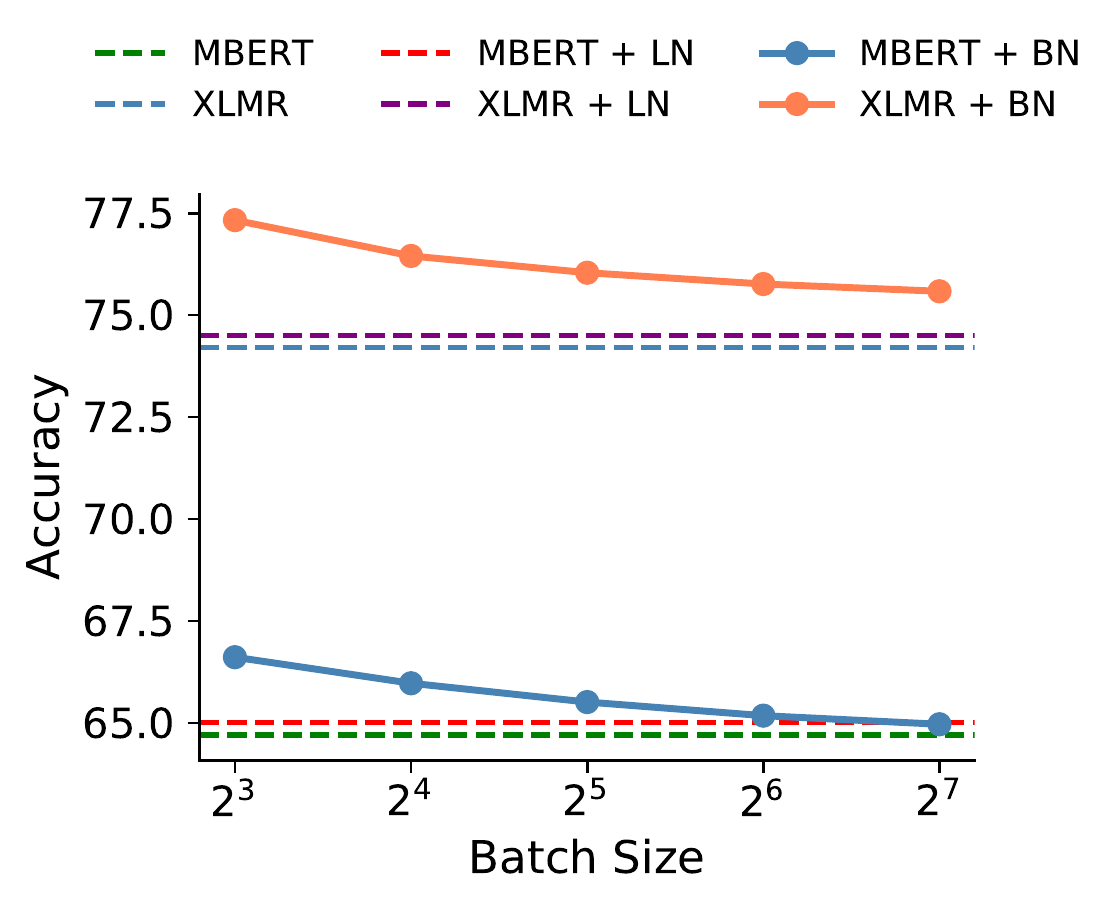}
 \caption{Results on \xnli{} on average across all language pairs. BN and LN denote batch and layer normalizations, respectively.
 }
 \label{fig:bn-ln}  
\end{figure} 
}

\newcommand{\insertLayerPlots}{
\begin{figure*}[tbh]
\centering
\begin{subfigure}[b]{0.245\linewidth}
    \centering
    \includegraphics[width=\textwidth]{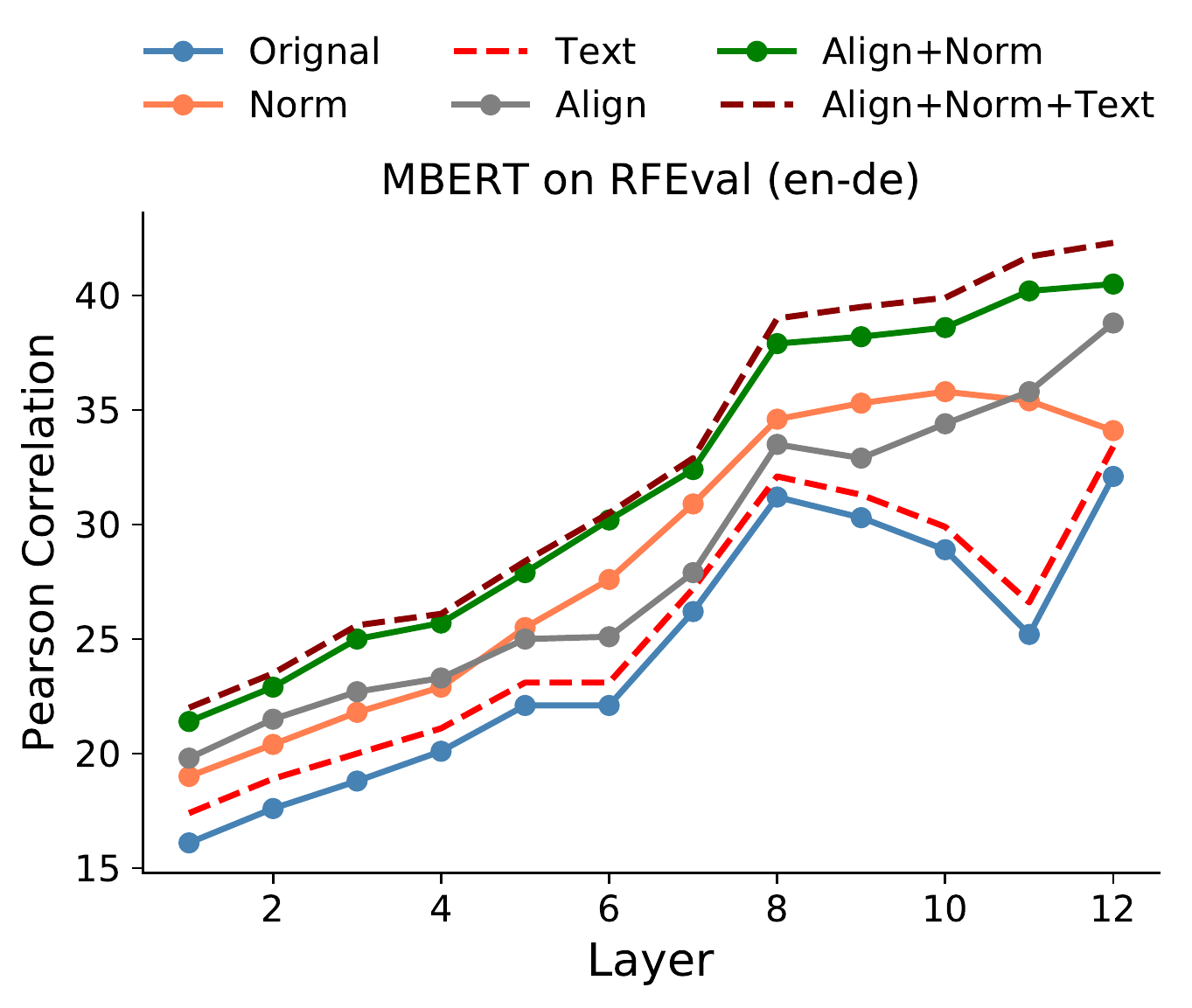}
\end{subfigure}
\hfill
\begin{subfigure}[b]{0.245\linewidth}  
    \centering 
    \includegraphics[width=\textwidth]{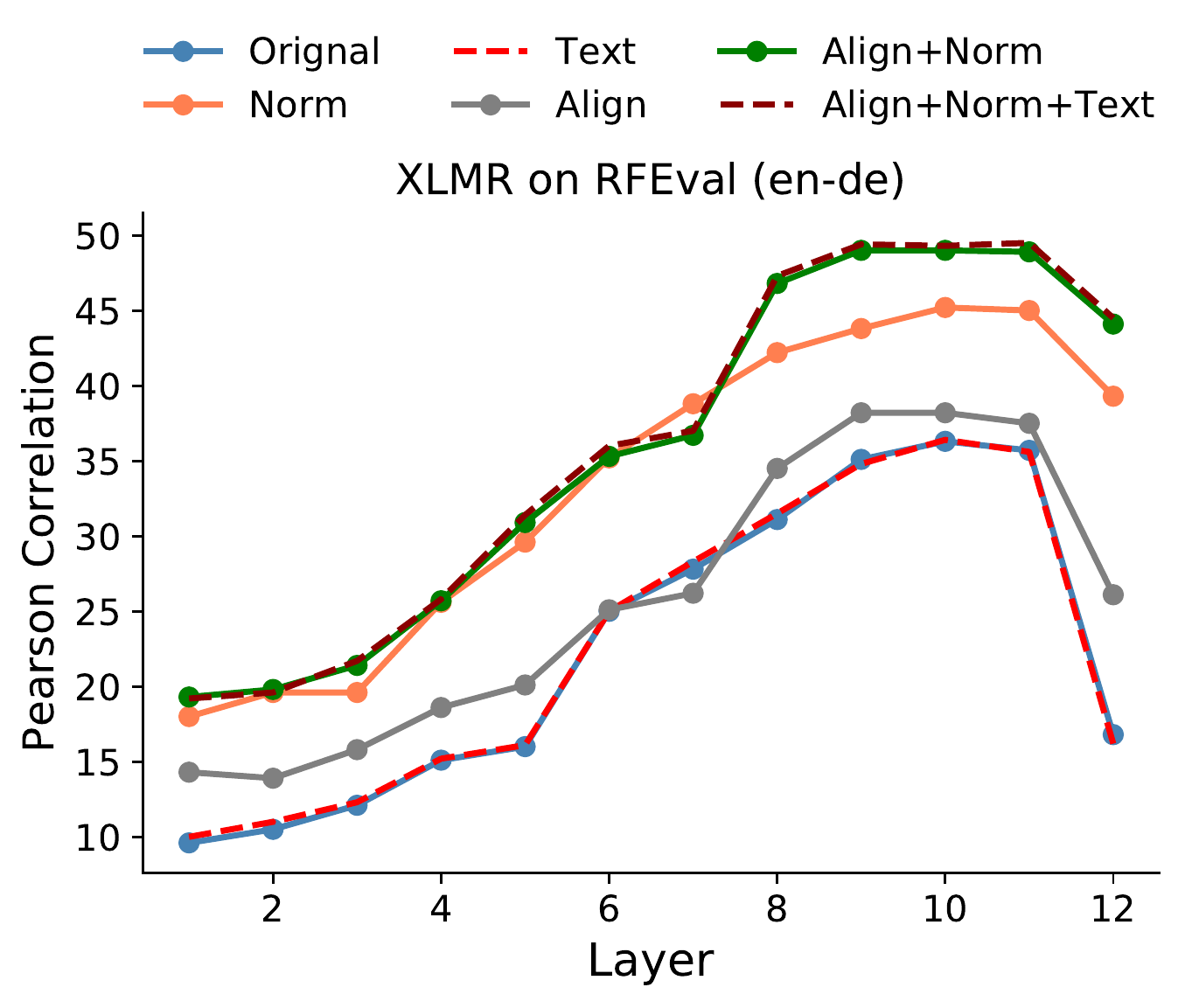}
\end{subfigure}
\hfill
\begin{subfigure}[b]{0.245\linewidth}  
    \centering 
    \includegraphics[width=\textwidth]{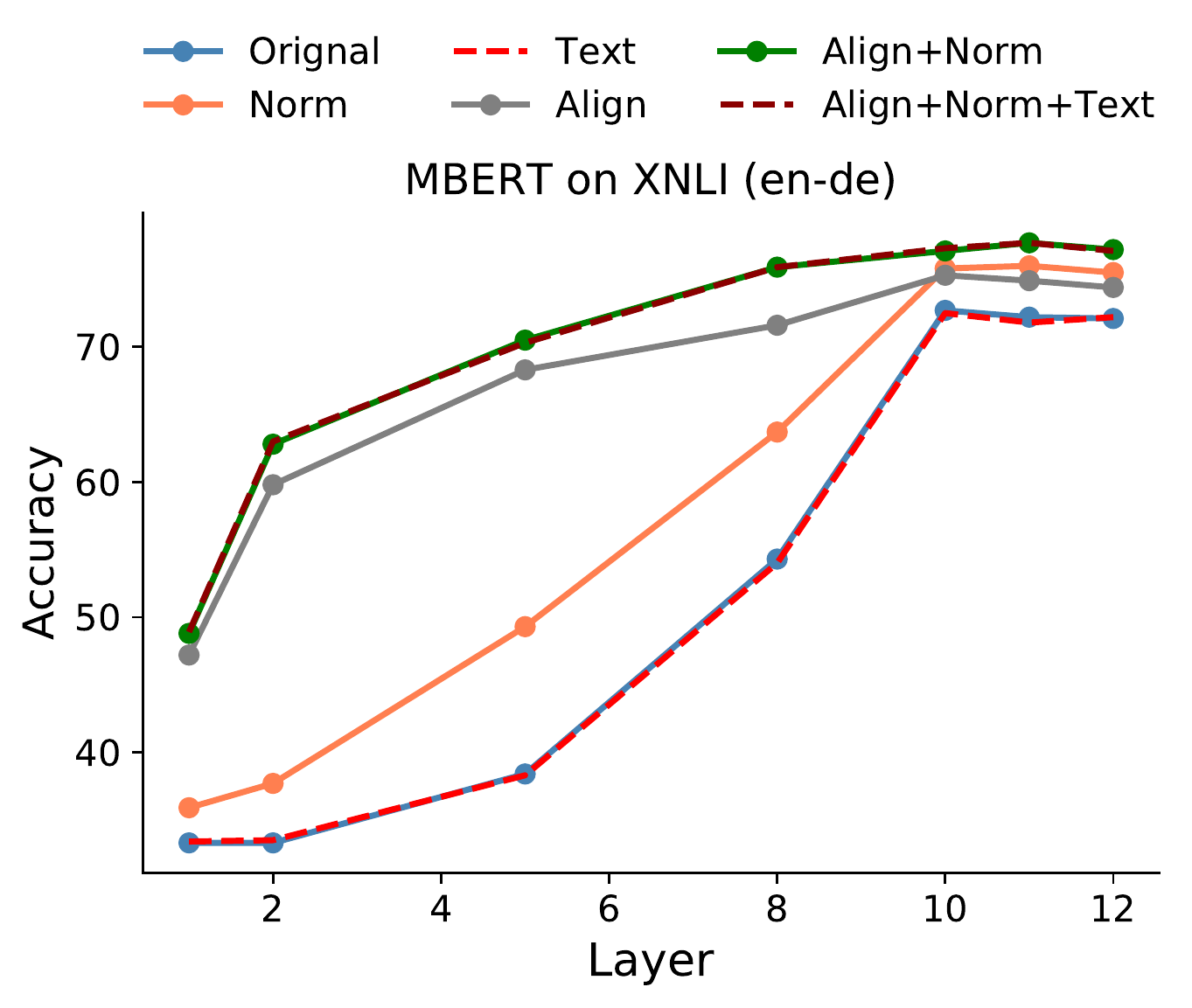}
\end{subfigure}
\hfill
\begin{subfigure}[b]{0.245\linewidth}  
    \centering 
    \includegraphics[width=\textwidth]{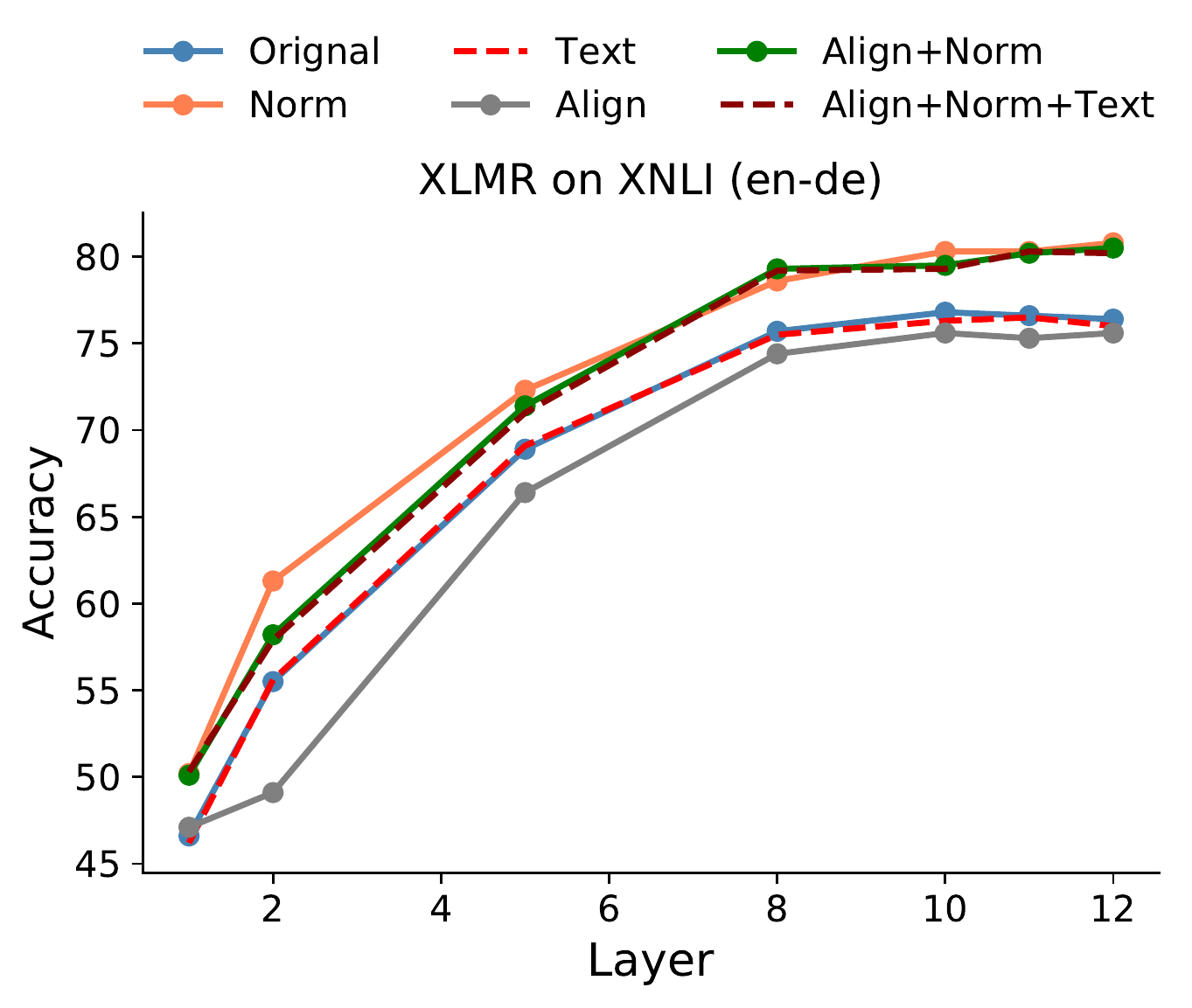}
\end{subfigure}

\caption{Results of m-BERT and XLM-R and our modifications across layers on the \rf{} and \xnli{} tasks. \label{fig:layers-de-en}} 
\end{figure*}
}
\newcommand{\insertLayerEnJv}{
\begin{figure}
\centering
\begin{subfigure}[b]{0.45\linewidth}
    \centering
    \includegraphics[width=\textwidth]{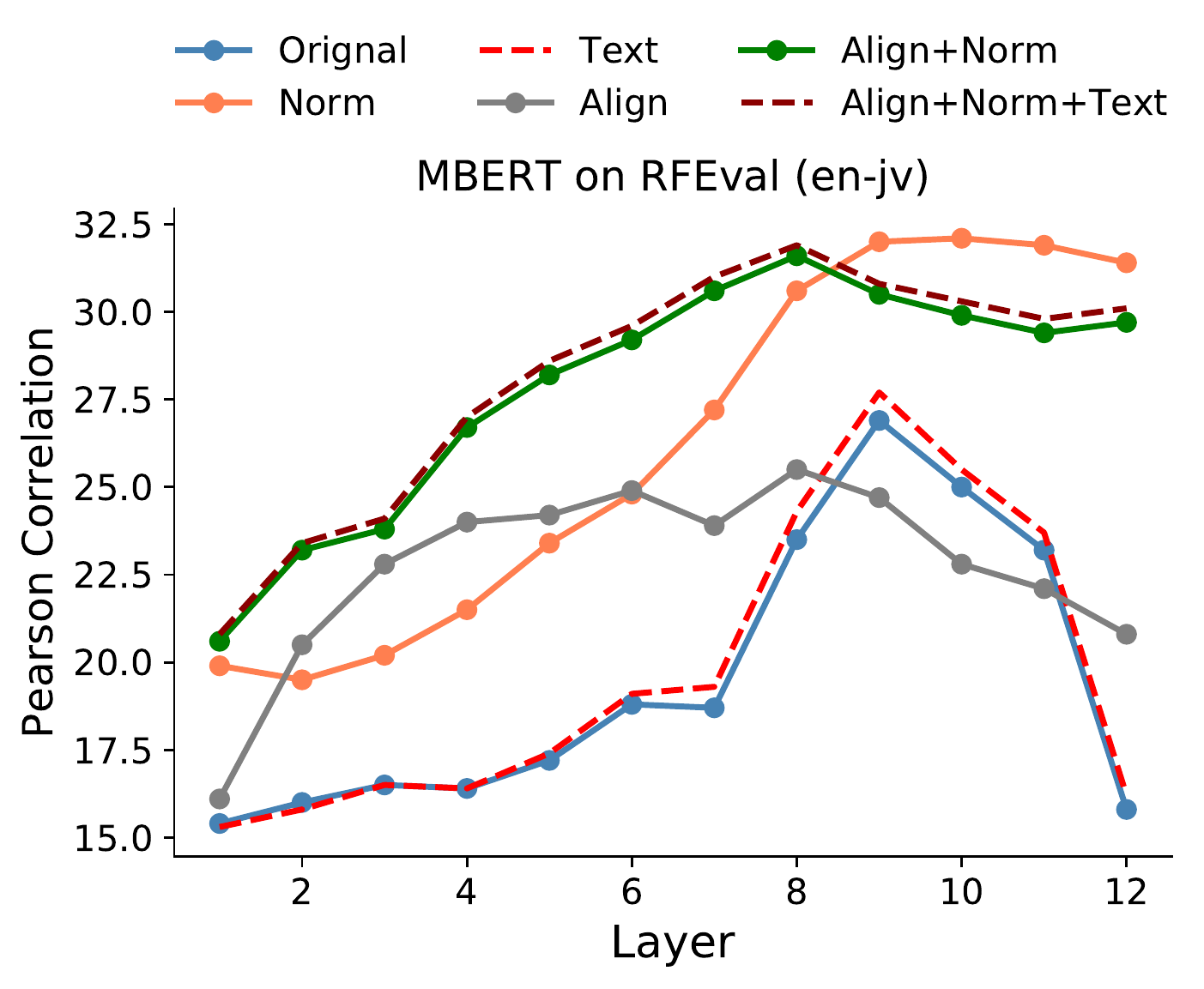}
\end{subfigure}
\hfill
\begin{subfigure}[b]{0.45\linewidth}  
    \centering 
    \includegraphics[width=\textwidth]{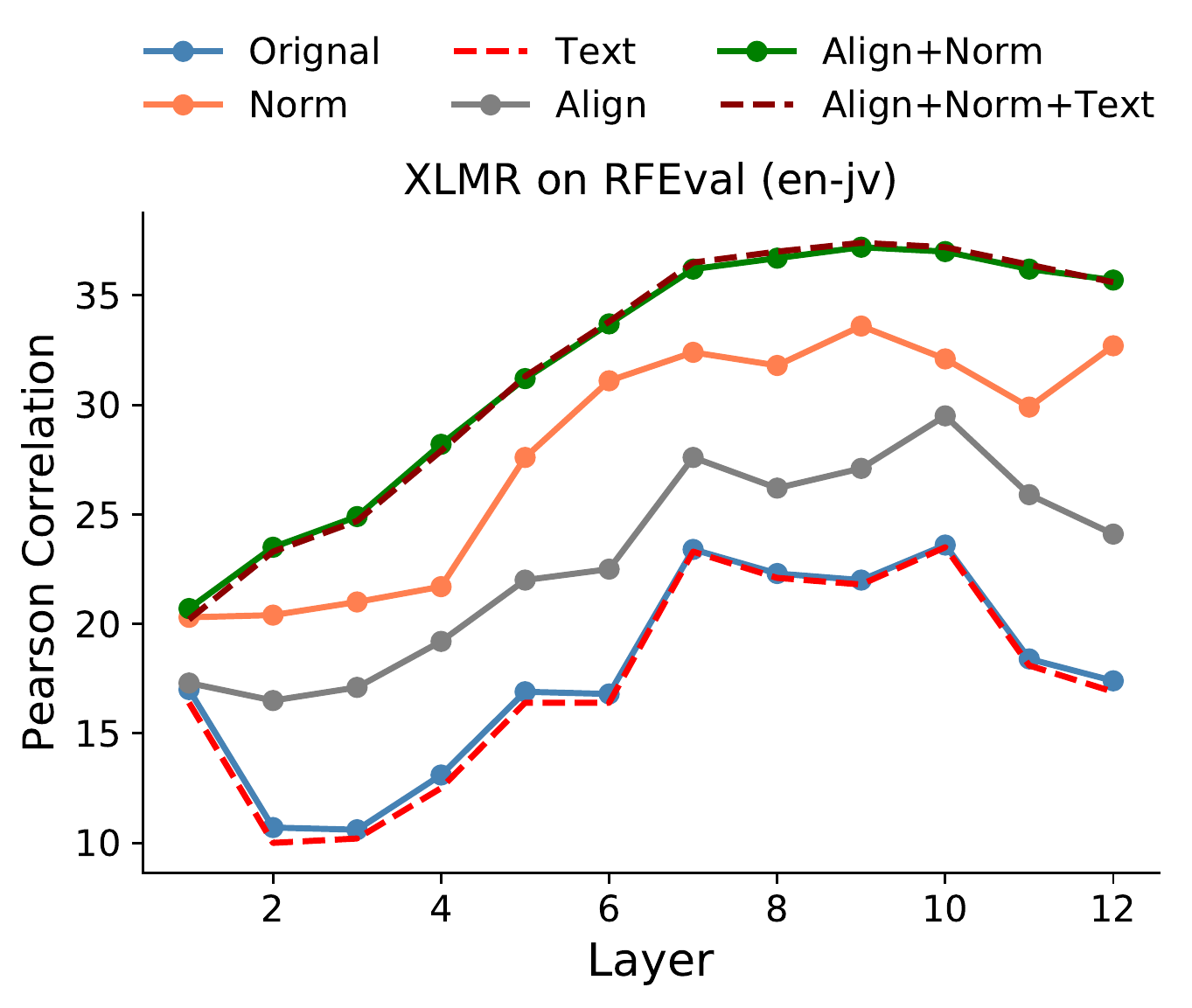}
\end{subfigure}

\caption{Results in one low-resource language pair (en-jv). \label{fig:en-jv-layers}} 
\end{figure}
}

\newcommand{\insertLayerTwoLangs}{
\begin{figure}
\centering
\begin{subfigure}[b]{0.45\linewidth}
    \centering
    \includegraphics[width=\textwidth]{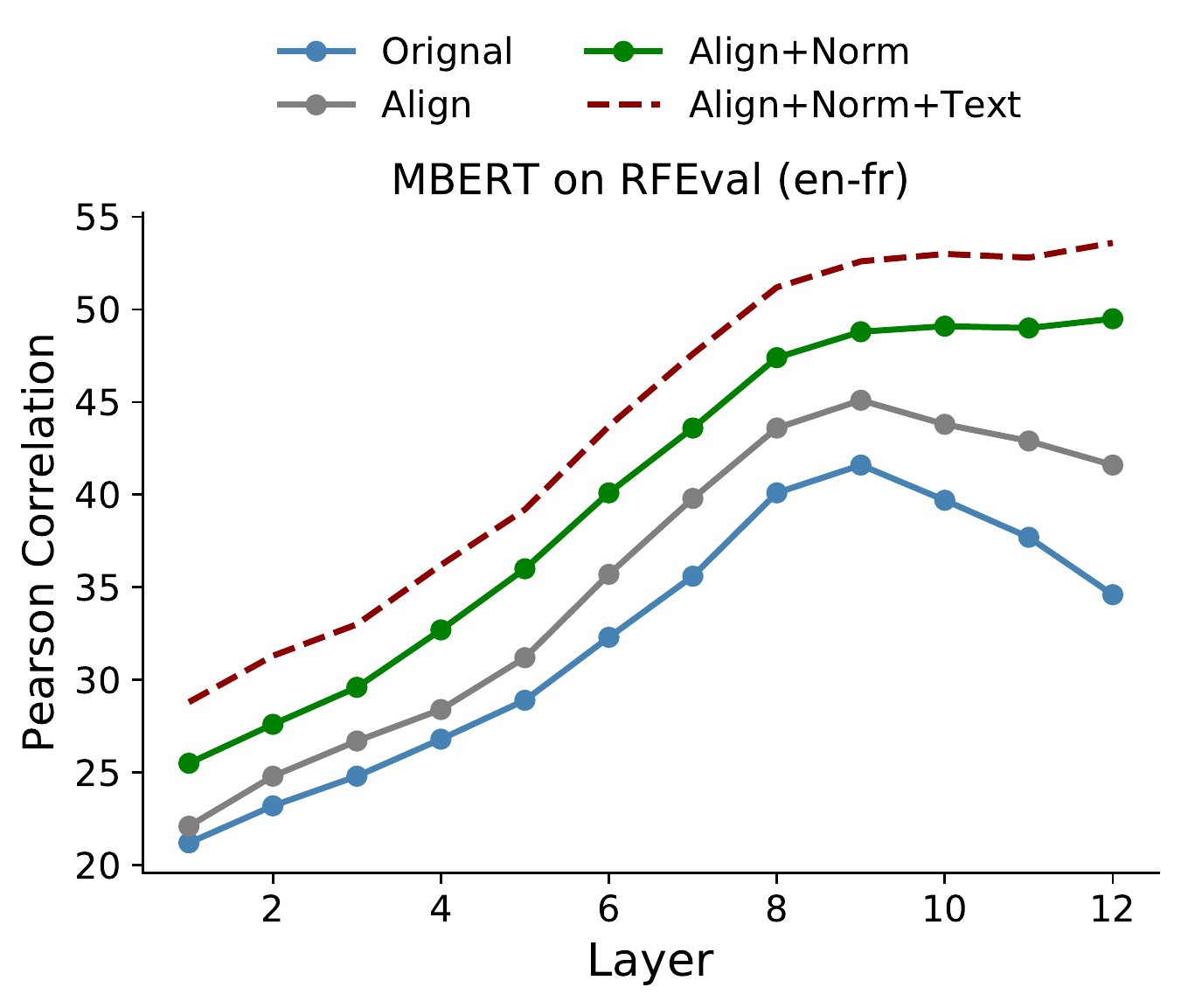}
\end{subfigure}
\hfill
\begin{subfigure}[b]{0.45\linewidth}  
    \centering 
    \includegraphics[width=\textwidth]{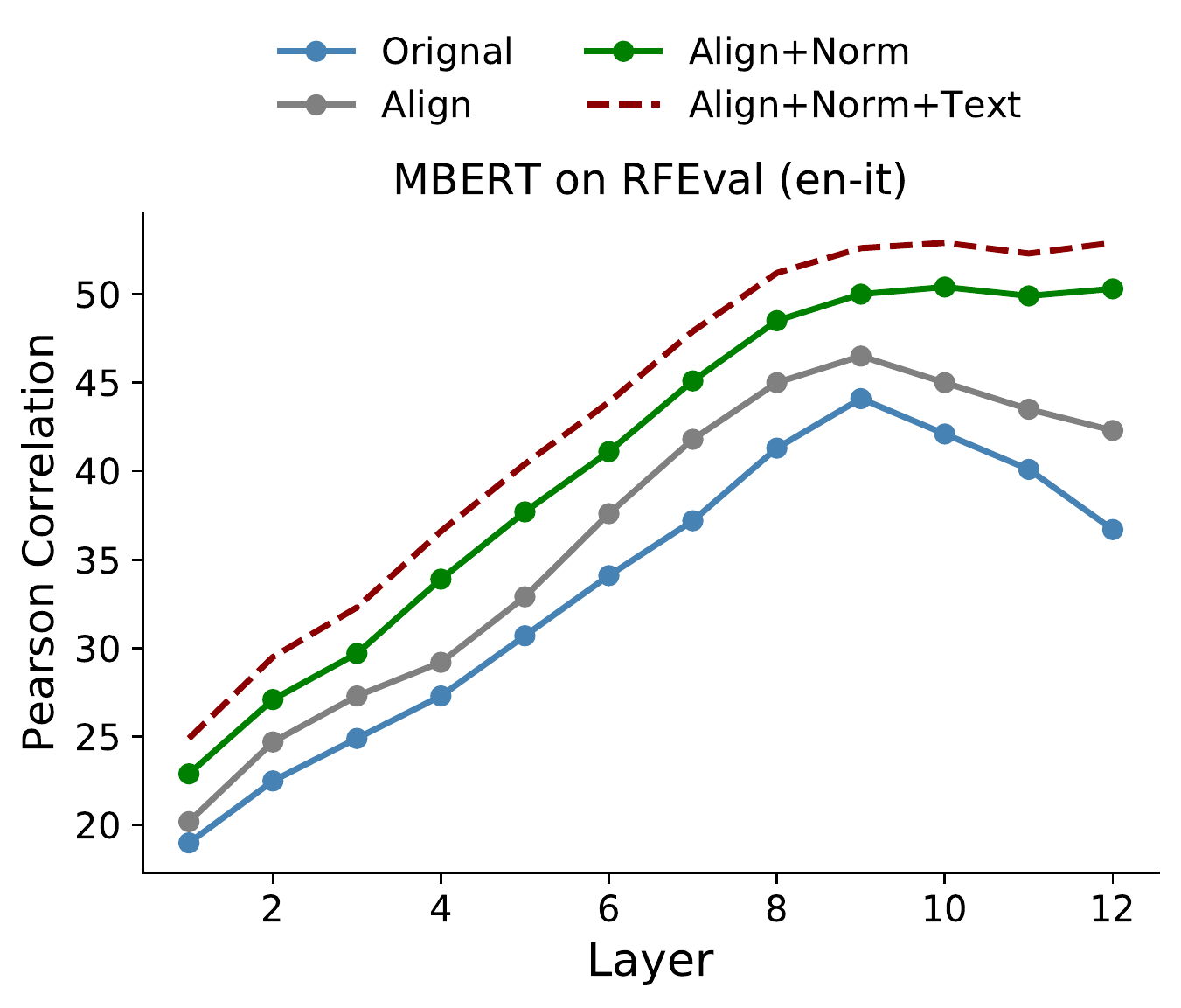}
\end{subfigure}

\caption{Results of m-BERT across layers on \rf{}. \label{fig:combination-across-layers}} 
\end{figure}
}

\newcommand{\insertCorrelationPlots}{
\begin{figure}[!tb]
\centering
\begin{subfigure}[b]{0.7\linewidth}
    \centering
    \includegraphics[width=\textwidth]{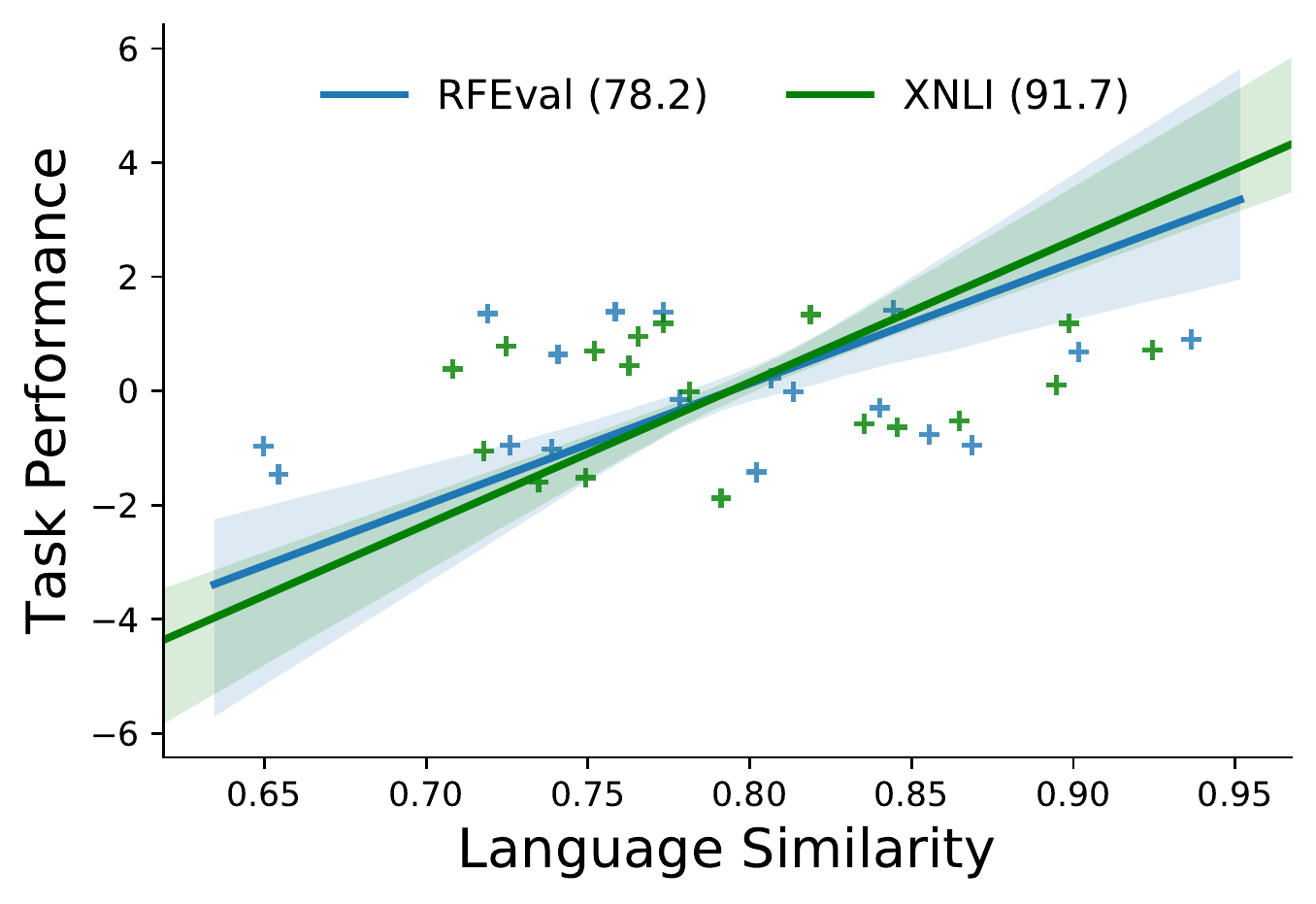}
\end{subfigure}
\hfill
\begin{subfigure}[b]{0.7\linewidth}  
    \centering 
    \includegraphics[width=\textwidth]{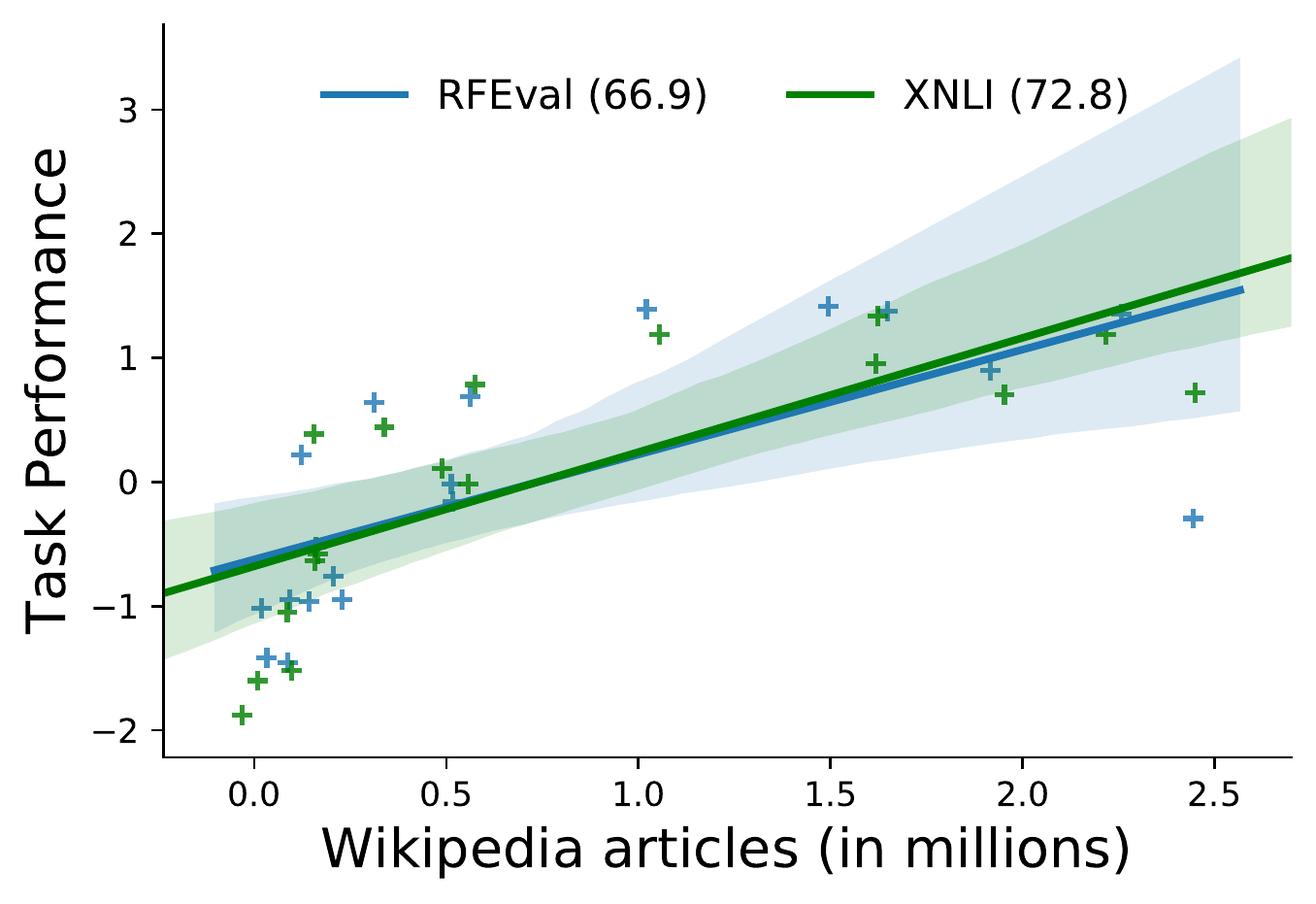}
\end{subfigure}
\caption{
Zero-shot performance on \xnli{} and \rf{} vs.\ 
language similarity to English (top), and data sizes in Wikipedia (bottom).
Each point is a language; brackets give the Pearson correlation of points on the x- and y-axis.
Zero-shot performance is based on the last layer of m-BERT and is standardized 
(zero mean, unit standard deviation) for better comparison.
 \label{fig:correlation}}
\end{figure}
}

\newcommand{\insertPerfGainsByText}{
\begin{figure*}[thb]
\centering
\begin{subfigure}[b]{0.3\linewidth}
    \centering
    \includegraphics[width=\textwidth]{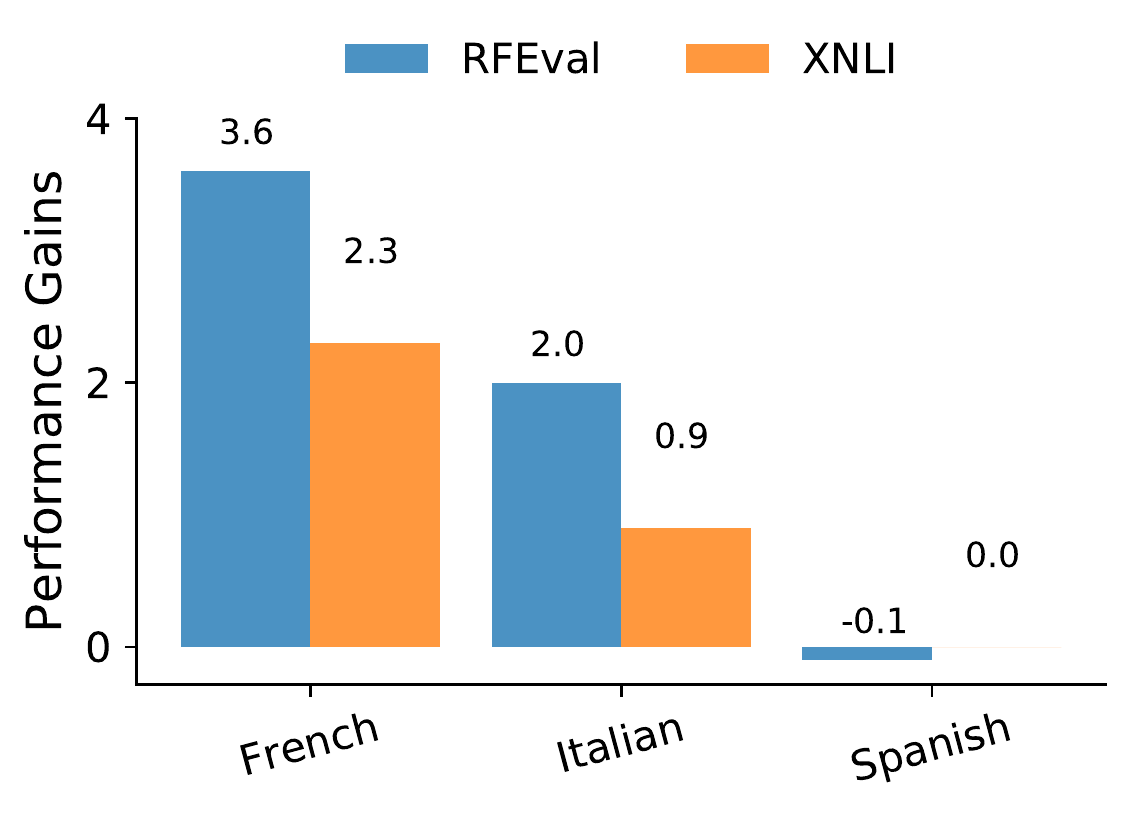}
\caption{Removing contractions}
\end{subfigure}
\hfill
\begin{subfigure}[b]{0.3\linewidth}  
    \centering 
    \includegraphics[width=\textwidth]{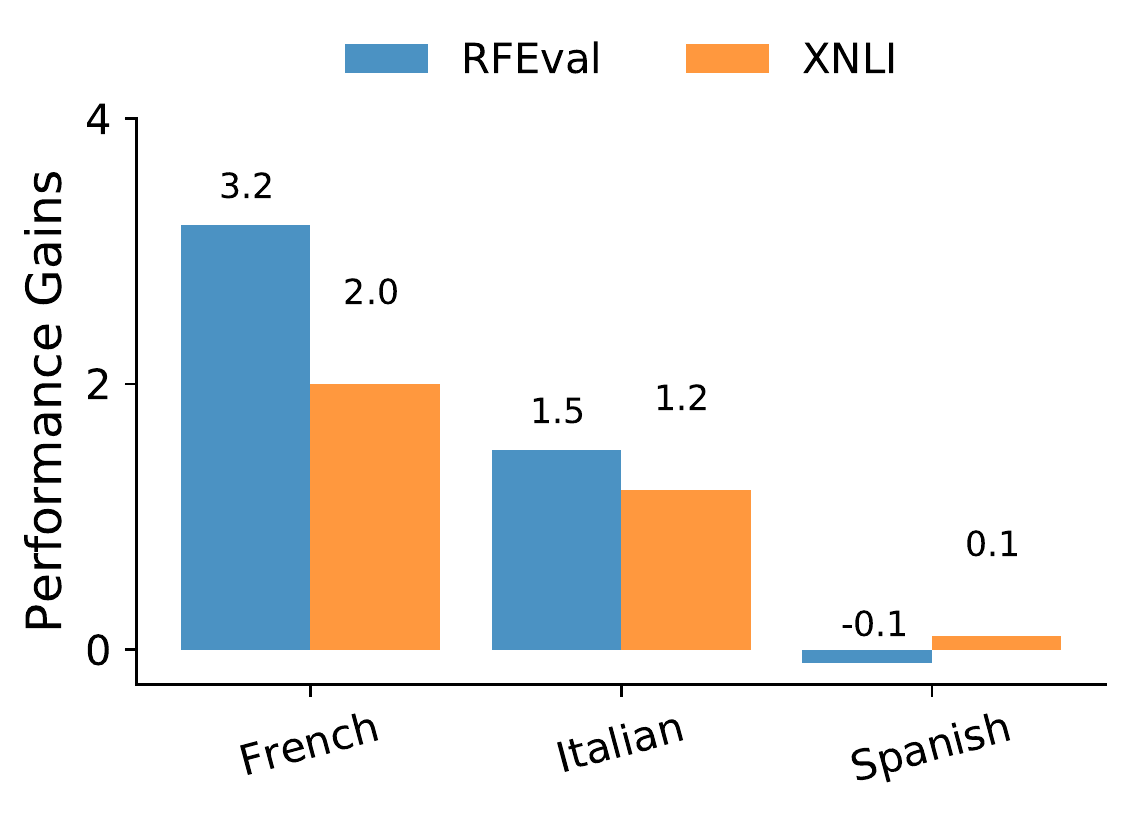}
\caption{Reversing adjective-noun order}    
\end{subfigure}
\hfill
\begin{subfigure}[b]{0.3\linewidth}  
    \centering 
    \includegraphics[width=\textwidth]{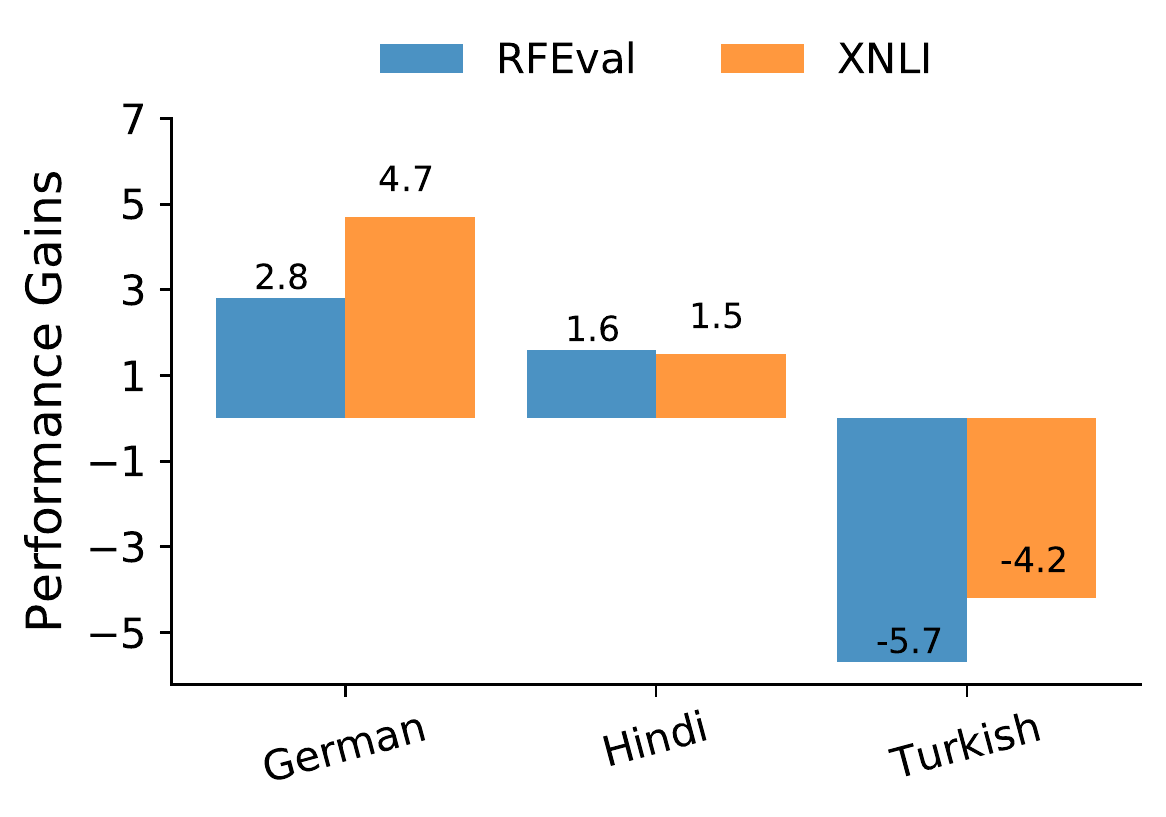}
\caption{Reversing object-verb order}
\end{subfigure}
\caption{Performance gains on \rf{} and \xnli{} obtained by three types of \metric{Text} \se{operations 
}. 
}

\label{fig:perf_gains_text}
\end{figure*} 
}
\newcommand{\insertLangsDes}{
\begin{table}[t]
    \setlength{\tabcolsep}{2.2pt}
    \footnotesize    
    \centering
    \begin{tabular}{@{}l l r c l l@{}}
    \toprule
    Language & \begin{tabular}[c]{@{}l@{}}Lang.\\ family\end{tabular} & \begin{tabular}[c]{@{}l@{}}Distance\\ (EN-X)\end{tabular} & \begin{tabular}[c]{@{}l@{}}Wiki-articles\\
    (in millions)\end{tabular} & \begin{tabular}[c]{@{}l@{}}Sim\\level\end{tabular} & \begin{tabular}[c]{@{}l@{}}Res\\level\end{tabular} \\
    \midrule
    Tagalog & $\alpha$ & 29.3 & 0.08 & low & low\\
    Javanese & $\alpha$ & 26.5 & 0.06 & low & low\\
    Bengali & $\gamma$ & 24.8 & 0.08 & low & low\\
    Marathi & $\gamma$ & 24.0 & 0.06 & low & low\\
    Estonian & $\eta$ & 23.8 & 0.20 & low & middle\\
    Hindi & $\gamma$ & 22.2 & 0.13 & middle & low\\
    Urdu & $\gamma$ & 21.7 & 0.15 & middle & middle\\
    Finnish & $\eta$ & 20.1 & 0.47 & middle & middle\\
    Hungarian & $\eta$ & 19.8 & 0.46 & middle & middle\\
    Afrikaans & $\beta$ & 19.6 & 0.09 & middle & low\\
    Malay & $\alpha$ & 19.2 & 0.33 & middle & middle\\
    Spanish & $\delta$ & 18.5 & 1.56 & high & high\\
    French & $\delta$ & 18.2 & 2.16 & high & high\\
    Italian & $\delta$ & 18.0 & 1.57 & high & high\\
    Indonesian & $\alpha$ & 17.7 & 0.51 & high & middle\\
    Dutch & $\beta$ & 16.3 & 1.99 & high & high\\
    Portuguese & $\delta$ & 16.2 & 1.02 & high & high\\
    German & $\beta$ & 15.6 & 2.37 & high & high\\
    English & $\beta$ & 0.0 & 5.98 & high & high\\
    \bottomrule
    \end{tabular}
    \caption{Languages used, with their language families: Austronesian ($\alpha$), Germanic ($\beta$), Indo-Aryan ($\gamma$), Romance ($\delta$), and Uralic ($\eta$). The cosine distances between target languages and English are
    measured using m-BERT. 
    }
    \label{tab:langs} 
\end{table}
}
\newcommand{\insertMainResults}{
\begin{table*}[h!]
\begin{subtable}{1\textwidth}
\sisetup{table-format=1.2} 
    \setlength{\tabcolsep}{2.8pt}
    \footnotesize    
    \centering
   \begin{tabular}{l rr|rr|rr|rr|rr|rr}
    \toprule
    & & \multicolumn{11}{c}{Language Families}\\
    Model & Avg &$\triangle$ & $\alpha$(4) & $\triangle$ &
    $\beta$(3) & $\triangle$ & $\gamma$(4)& $\triangle$ & $\delta$(4) & $\triangle$ & $\eta$(3) & $\triangle$ \\
    \midrule
    \multicolumn{7}{l}{\textit{Original cross-lingual embeddings}}\\
    \metric{m-BERT} & 38.0 &-& 36.6&-& 40.4&-& 28.2&-& 49.8&-& 34.8&-\\
    \metric{XLM-R} & 12.9 &-& 13.5&-& 17.4&-& 2.9&-& 25.9&-& 11.6&-\\
    \midrule
    \multicolumn{7}{l}{\textit{Modified cross-lingual embeddings}}\\
    \metric{m-BERT $\oplus$ Joint-Align $\oplus$ Norm} & 48.1 &$+$10.1 & 45.9 &$+$9.3&  47.5 &$+$7.1 & 32.4 & $+$4.2 & 53.4 &$+$3.6 & 46.0 &$+$11.2\\
    \metric{XLM-R $\oplus$ Joint-Align $\oplus$ Norm} & 46.4 &$+$33.5 & 46.5 &$+$33.0& 48.2 &$+$30.8& 37.0 &$+$34.1& 53.8 &$+$27.9& 47.2 &$+$35.6\\
	\bottomrule  
    \end{tabular}
   \caption{Cross-lingual Semantic Text Similarity on the \rf{} task}\label{tab:sub_third}
\end{subtable}

\bigskip
\begin{subtable}{1\textwidth}
\sisetup{table-format=1.2}
    \setlength{\tabcolsep}{3.6pt}
    \footnotesize    
    \centering
    \begin{tabular}{l rr|rr|rr|rr|rr|rr}
    \toprule
    & & \multicolumn{11}{c}{Language Families}\\
    Model & Avg &$\triangle$ & $\alpha$(4) & $\triangle$ &
    $\beta$(3) & $\triangle$ & $\gamma$(4)& $\triangle$ & $\delta$(4) & $\triangle$ & $\eta$(3) & $\triangle$ \\
    \midrule
    \multicolumn{13}{l}{\textit{Original cross-lingual embeddings}}\\
    \metric{m-BERT} & 64.7 &-& 60.8 &-& 69.1 &-& 57.9 &-& 73.1 &-& 63.4 &-\\
    \metric{XLM-R} & 74.8 &-& 72.4 &-& 76.3 &-& 70.9 &-& 78.4 &-& 76.1 &-\\
    \midrule
    \multicolumn{13}{l}{\textit{Modified cross-lingual embeddings}}\\
    \metric{m-BERT $\oplus$ Joint-Align $\oplus$ Norm} & 72.3 & $+$7.6 & 72.3 &$+$11.5& 75.8 &$+$6.7& 65.2 &$+$7.3& 77.4 &$+$4.3& 72.0 &$+$8.6\\
    \metric{XLM-R $\oplus$ Joint-Align $\oplus$ Norm} & 77.6 & $+$2.8 & 74.8 &$+$2.4& 79.6 &$+$3.3& 73.7 &$+$2.8& 80.9 &$+$2.5& 78.8 &$+$2.7\\
	\bottomrule  
    \end{tabular}
   \caption{Cross-lingual Zero-shot transfer on the \xnli{} task}\label{tab:sub_third}
   
\end{subtable}
\caption{Overall results of established cross-lingual baselines and our modifications, for \rf{} and \xnli{}. Brackets denote the number of languages per group. Results are averaged per group. $\triangle$ is the difference between the performance of the original and the modified encoders.} \label{tab:main_results}
\end{table*}
}

\newcommand{\insertWALS}{
\begin{table}
    \setlength{\tabcolsep}{4pt}
    \footnotesize    
    \centering
    \begin{tabular}{l ccc}
    \toprule
    Model & $\tau$ & $r$ & $\rho$ \\
    \midrule
    \metric{m-BERT} & 53.2 & 74.7 & 71.8 \\
    \metric{XLM-R} & 54.4 & 70.1 & 73.5 \\
    \metric{m-BERT $\oplus$ Joint-Align $\oplus$ Norm} & 17.5 & 57.3 & 21.2 \\
    \metric{XLM-R $\oplus$ Joint-Align $\oplus$ Norm} & 15.9 & 57.7 & 26.0\\
	\bottomrule  
    \end{tabular}
    \caption{Correlations (Kendall $\tau$, Pearson $r$ and Spearman $\rho$) between language similarities induced by m-BERT/XLM-R and WALS for 19 languages. 
    \label{tab:wals}}
\end{table}
}

\newcommand{\insertPerfGaps}{
\begin{table}
    \setlength{\tabcolsep}{2.4pt}
    \footnotesize    
    \centering
    \begin{tabular}{l rrr}
    \toprule
    Model & \xnli{} & \rf{} & Avg\\
    \midrule
    \metric{m-BERT} & 17.4 & 24.5 & 21.0 \\
    \metric{XLM-R} & 11.1 & 37.8 & 24.5 \\
    \metric{m-BERT $\oplus$ Joint-Align $\oplus$ Norm} & 9.8 & 14.4 & 12.1 \\
    \metric{XLM-R $\oplus$ Joint-Align $\oplus$ Norm} & 8.4 & 4.3 & 6.3 \\
	\bottomrule  
    \end{tabular}
    \caption{Performance gap (lower is better) for cross-lingual classification transfer, and reference-based and reference-free MT.
    \label{tab:perf_gaps}}
\end{table}
}

\newcommand{\insertAblation}{
\begin{table}[h!]
    \setlength{\tabcolsep}{4pt}
    \footnotesize    
    \centering
    \begin{tabular}{l rr}
    \toprule
    Model & \xnli{} & \rf{} \\
    \midrule
    \metric{m-BERT $\oplus$ Norm} & $+$1.9 & $+$1.7 \\
    \metric{m-BERT $\oplus$ Joint-Align} & $+$5.2 & $+$7.6  \\
    \metric{m-BERT $\oplus$ Joint-Align $\oplus$ Norm} & $+$7.6 & $+$10.1 \\
    \metric{XLM-R $\oplus$ Norm} & $+$2.5 & $+$27.1 \\
    \metric{XLM-R $\oplus$ Joint-Align} & $-$0.2 & $+$11.6 \\
    \metric{XLM-R $\oplus$ Joint-Align $\oplus$ Norm} & $+$2.8 & $+$33.5 \\
	\bottomrule  
    \end{tabular}
    \caption{Ablation tests of our modified encoders. Performance gains are averaged over all languages.
    \label{tab:ablation}}
\end{table}
}

\aclfinalcopy 

\newcommand{\se}[1]{\textcolor{black}{#1}}
\newcommand{\wei}[1]{\textcolor{black}{#1}}

\newcommand{\sem}[1]{\textcolor{black}{#1}}
\newcommand{\rf}[1]{\textsf{RFEval}}
\newcommand{\xnli}[1]{\textsf{XNLI}}

\title{Inducing Language-Agnostic Multilingual Representations}

\author{Wei Zhao$^{\dagger}$ \text{ } Steffen Eger$^{\dagger}$ \text{ }
Johannes Bjerva$^{\Psi,\Phi}$ \text{ } Isabelle Augenstein$^{\Phi}$\\
    $^\dagger$Technische Universit\"at Darmstadt 
    \text{ } $^{\Phi}$University of Copenhagen 
    \text{ } $^{\Psi}$Aalborg University \\
    {\tt \{zhao,eger\}@aiphes.tu-darmstadt.de}\\
    {\tt
    jbjerva@cs.aau.dk}\\
    {\tt augenstein@di.ku.dk}
  }

\date{}

\begin{document}
\maketitle
\todo{SE: I think the two parts of the title don't fit together this way. Reversing the order would at least be better}
\begin{abstract}
Cross-lingual representations have the potential to 
make NLP techniques  
available to the vast majority of languages in the world. 
However, they currently require large pretraining corpora or 
access to typologically similar languages.
In this work, we address these 
obstacles by removing language identity signals from multilingual embeddings.
We examine three approaches for this: (i) re-aligning the vector spaces of target languages (all together) to a pivot source language; (ii) removing language-specific means and variances,
which yields better discriminativeness of embeddings as a by-product; and (iii) increasing input similarity across languages by removing morphological contractions and sentence reordering. 
We evaluate 
on \xnli{} and reference-free MT 
across 19 typologically diverse languages. 
Our findings expose the limitations of
these approaches---unlike vector normalization, vector space re-alignment and text normalization do not achieve consistent gains across encoders and languages. 
Due to the approaches' additive effects, their combination  
decreases the cross-lingual transfer gap by 8.9 points (m-BERT) and 18.2 points (XLM-R) on average across all tasks and languages, however. Our code and models are publicly available.\footnote{\url{https://github.com/AIPHES/Language-Agnostic-Contextualized-Encoders}}
\end{abstract}

\section{Introduction}

Cross-lingual 
text representations \cite{devlin:2019,Conneau:2019a} 
ideally allow for
transfer between \textit{any} language pair, 
and thus hold the promise to
alleviate the data sparsity problem for low-resource languages. However, 
until now, cross-lingual systems trained on English appear to 
transfer poorly to target languages dissimilar to English \citep{wu:2019-beto,pires:2019} and for which only small monolingual corpora are available \citep{Conneau:2019a,Hu:2020,Lauscher:2020}, as illustrated in Fig.~\ref{fig:correlation}.\footnote{\wei{We consider language similarity as the cosine similarity between the average representations of two languages over monolingual corpora from Wikipedia.}}  
\insertCorrelationPlots

As a remedy, recent work has suggested to train representations on larger multilingual corpora \citep{Conneau:2019a} and, more importantly, to re-align them post-hoc so as to address the deficits of state-of-the-art contextualized encoders which have not seen any parallel data during training \citep{schuster:2019,wu:2019-beto,Cao:2020}. 
However, re-mapping (i) can be costly, (ii) requires parallel data on word or sentence level, which may not be available abundantly in low-resource settings, and (iii) its positive effect has not yet been studied systematically.  

Here, we explore 
\emph{normalization} as an alternative to re-mapping. To decrease the distance between languages and thus allow for better cross-lingual transfer, we normalize (i) text inputs to encoders before vectorization to increase cross-lingual similarity, \wei{e.g., by reordering sentences according to typological features,
}
and (ii) the representations themselves by removing their means and standard deviations, a common operation in machine and deep learning \citep{LeCun:1998,rueckle:2018}. 
We evaluate vector normalization and post-hoc re-mapping across a typologically diverse set of 19 languages from five language families with varying sizes of  monolingual corpora. However, input normalization is examined on a smaller sample of languages, as it is not feasible for languages whose linguistic features cannot be obtained automatically.
We investigate two NLP tasks, and two state-of-the-art contextualized cross-lingual encoders---multilingual BERT \cite{devlin:2019} and XLM-R \cite{Conneau:2019a}. 
Further, we provide a thorough analysis to investigate the effects of these techniques: (1) across layers; (2) to decrease the cross-lingual transfer gap, especially for low-resource and dissimilar languages; and (3) 
to eliminate language identity signals from multilingual representations and thus induce language-agnostic representations.

We evaluate on two cross-lingual tasks of varying difficulty: 
(1) zero-shot cross-lingual 
natural language inference (\xnli{}) 
measures the transfer ability of inference from source to target languages, where only the source language is annotated;
and (2) reference-free machine translation evaluation (\rf{}) measures the ability of multilingual embeddings to assign adequate cross-lingual semantic similarity scores to text from two languages, where one is frequently a corrupt automatic translation. 

\textbf{Our contributions}: We show that: 
(i) 
 input normalization leads to  performance gains of up to 4.7 points on two challenging tasks; 
(ii) 
normalizing vector spaces is surprisingly effective, rivals much more resource-intensive methods such as re-mapping, and leads to more consistent gains; 
(iii) all three techniques---vector space normalization,  re-mapping and input normalization---are orthogonal and their gains often stack. 
This is a very important finding as it allows for improvements on a much larger scale,  
especially 
for typologically dissimilar and low-resource languages. 

\section{Related Work}


\paragraph{Cross-lingual Transfer}
Static cross-lingual representations have long been used for effective cross-lingual transfer and can even be induced without parallel data  \citep{artetxe-etal-2017-learning,lample2018word}. 
In the monolingual case, static cross-lingual embeddings have recently been succeeded by contextualized ones, which yield considerably better results. 
The capabilities and limitations of the contextualized multilingual BERT 
(m-BERT) representations is a topic of vivid discourse. 
\citet{pires:2019} show surprisingly good transfer performance for m-BERT despite it being trained without parallel data, and that transfer is better for typologically similar languages. 
\citet{Wu:2019} show that language representations are not correctly aligned in 
m-BERT, 
but can be linearly re-mapped. Extending this, \citet{Cao:2020} find that jointly aligning language representations to be more useful than language-independent rotations. However, we show that the discriminativeness of the resulting embeddings is still poor, i.e., 
random word pairs are often assigned very high cosine similarity scores by the upper layers of original encoders, 
especially for 
XLM-R. 

\citet{Libovick:2019} further observe that m-BERT representations of related languages 
are seemingly close to one another in the cross-lingual embedding space. 
They show that removing 
language-specific means from m-BERT 
can 
eliminate language identity signals.
In contrast, we remove both language-specific means and variances as well as morphological contractions, and reorder sentences to reduce linguistic gaps between languages. 
In addition, 
our analysis covers more languages from a typologically broader sample, and shows that vector space normalization is as effective as other recently proposed fixes for m-BERT's limitations (especially re-mapping), but is much cheaper and orthogonal to other solutions (e.g., input normalization) in that gains are almost additive.
\fi

\paragraph{Linguistic Typology in NLP.}
\insertDistplots
Structural properties of many of the world's languages can be queried via databases such as WALS
\citep{wals}. 
\citet{o2016survey,ponti2019modeling} suggest to inject typological information into models to bridge the performance gap between high- and low-resource languages.
\citet{bjerva_augenstein:2018,delhoneux2018parameter,bjerva-augenstein-2021-typological} show that cross-lingual transfer can be more successful between languages which share, e.g., morphological properties. 
We draw inspiration from \citet{galactic_dependencies}, who use dependency statistics to generate a large collection of synthetic languages to augment training data for low-resource languages.
This intuition of modifying languages based on syntactic features can also be used in order to decrease syntactic and morphological differences between languages.
We go further than using syntactic features, and remove word contractions and reorder sentences based on typological information from WALS.

\section{Language-Agnostic Representations}
Analyses by \citet{ethayarajh:2019} indicate that random words are often assigned high cosine similarities in the upper layers of monolingual BERT. 
We examine this in a cross-lingual setting,
by randomly 
selecting 500 German-English mutual word translations and random word pairs within parallel sentences from Europarl~\citep{koehn:2005}. 
\wei{Fig.~\ref{fig:histogram} (left) shows 
histograms 
based on the last layers of m-BERT \citep{devlin:2019} and XLM-R \citep{Conneau:2019a}, respectively, which show that XLM-R wrongly assigns nearly perfect cosine similarity scores (+1) to both mutual word translations (matched word pairs) and random word pairs, whereas m-BERT sometimes assigns low scores to mutual translations. 
This reaffirms that both m-BERT and XLM-R 
have difficulty in distinguishing matched from random word pairs. 
Surprisingly, vector space re-mapping does not seem to help for XLM-R, but better separates random from matched pairs for m-BERT (Fig.~\ref{fig:histogram} (middle)). In contrast, the joint effect of normalization and re-mapping leads to adequate separation of the two distributions for both m-BERT and XLM-R, increasing the discriminative ability of both encoders.
}

\subsection{Vector space re-alignment}
\label{sec:align}
m-BERT and XLM-R induce cross-lingual vector spaces in an unsupervised way---no parallel data is involved at training time. To improve upon these representations, recent work has suggested to re-map them, 
i.e., to use small amounts of parallel data to restructure the cross-lingual vector spaces.  
We follow the joint re-mapping approach of \citet{Cao:2020}, which has shown better results than rotation-based 
re-mapping. 


\paragraph{Notation.} 
Suppose we have $k$ parallel corpora 
$C^1,\dots, C^k$, \se{i.e.}, 
$C^{\nu}=\{(\mathbf{s}^1, \mathbf{t}^1), \dots, (\mathbf{s}^n,\mathbf{t}^n)\}$ is a set of \se{corresponding} sentence pairs from source and target languages, \se{for $\nu=1,\ldots,k$}. 
\se{We denote the alignments of words in a sentence pair} 
$(\mathbf{s},\mathbf{t})$ 
\se{as} 
$a(\mathbf{s},\mathbf{t})=\{(i_1, j_1), \dots, (i_m, j_m)\}$, where $(i,j)$ denotes 
\se{that} 
$\mathbf{s}_i$ and $\mathbf{s}_j$ are mutual translations. Let $f(i, \mathbf{u})$ be the contextual embedding for the $i$-th word in a sentence $\mathbf{u}$.

\paragraph{Joint Alignment via Fine-tuning.} 
We align the monolingual sub-spaces of a source and target language by minimizing the distances of embeddings for matched word pairs in the corpus $C^{\nu}$:
\begin{equation}\label{eqn:rotate}
\begin{split}
L(&C^{\nu}, f_{\Theta})\\
&=\sum_{(\mathbf{s}, \mathbf{t}) \in C^{\nu}} \sum_{(i, j) \in a(\mathbf{s}, \mathbf{t})}
\|f_{\Theta}(i, \mathbf{s}) - f_{\Theta}(j, \mathbf{t})) \|_2^2\quad
\end{split}
\end{equation}
where $\Theta$ are the parameters of the encoder $f$. As in \citet{Cao:2020}, we use a regularization term to avoid for the resulting \se{(re-aligned)} embeddings to drift \se{too far} away from the initial encoder state $f_0$: 
\begin{equation}\label{eqn:rotate}
R(C^{\nu},f_{\Theta})=\sum_{\mathbf{t} \in C^{\nu}}\sum_{i=1}^{\rm{len}(\mathbf{t})} \|f_{\Theta}(i, \mathbf{t}) - f_0(i, \mathbf{t})\|_2^2
\end{equation}
Like \se{for the} multilingual pre-training of m-BERT \se{and XLM-R}, we fine-tune the encoder $f$ on the concatenation of $k$ parallel corpora to handle resource-lean languages, which is in contrast to offline alignment with language-independent rotations \citep{aldarmaki:2019,schuster:2019}. 
\se{Assume that English is a common pivot (source language) in all our $k$ parallel corpora.}
Then the following objective function orients all non-English embeddings toward English:
\begin{equation}\label{eqn:rotate}
\min_{\Theta} \sum_{\nu=1}^{k}L(C^\nu,f_{\Theta}) + R(C^\nu,f_{\Theta})
\end{equation}

In \S\ref{sec:results}, we refer to the above described re-alignment step as \metric{Joint-Align}.

\subsection{Vector space normalization} 
\label{sec:norm}
We add a batch normalization layer that constrains all embeddings of different languages into a 
distribution 
with zero mean and unit variance:
\begin{equation}
    \bar{f}(i,\mathbf{s}) = \frac{f(i,\mathbf{s}) - \mu_{\beta}}{\sqrt{\sigma_{\beta}^2 + \epsilon}}
\end{equation}
where $\epsilon$ is a constant value for numerical stability, $\mu_\beta$ and $\sigma_\beta$ are \sem{ mean and variance, serving as per batch statistics for each time step in a sequence.}\todo{SE: What's a moving mean?}
In addition to a common effect during training, i.e., reducing covariate shift of input spaces, this additional layer in the cross-lingual setup may allow for 1) 
removing language identity signals, e.g. language-specific means and variances, from multilingual embeddings;
and 2) increasing the discriminativeness of embeddings 
so that they 
can distinguish 
word pairs with different senses,
as shown in Fig. \ref{fig:histogram} (right). 
%
\sem{We apply batch normalization to the last layer representations of m-BERT and XLM-R, and use a batch size of 8 across all setups.}
In \S\ref{sec:results}, we refer to the above batch normalization step as \metric{Norm} \sem{and contrast this with layer normalization. The latter yields batch-independent statistics, which are \todo{SE: "viz." is synonymous to "namely", so I think it's not correctly used here} computed across all time steps for individual input sequences in a batch. This is predominantly used to stabilize the training process of RNN~\citep{layernorm:2016} and Transformer-based models~\citep{transformer:2017}.} 




\subsection{Input normalization} 
In addition to joint alignment and vector space normalization, 
we investigate 
decreasing cross-linguistic differences between languages via the following surface form manipulation of input texts.

\paragraph{Removing Morphological Contractions.} 
In many languages, e.g.~Italian, prepositions and definite articles are often contracted.
For instance, \textit{de il} (\textit{`of the'}) is usually contracted to \textit{del}.
This leads to a mismatch between, e.g., English and Italian in terms of token alignments, and increases the cross-lingual difference between the two. We segment an orthographic token (e.g.~\textit{del})  into several (syntactic) tokens (e.g.~\textit{de il}).\footnote{We use UDPipe \citep{straka:2016}, which is a pipeline trained on UD treebank 2.5 \citep{Nivre:2020}.}
This yields a new sentence which no longer corresponds to typical standard Italian grammar, but which we hypothesise reduces the linguistic gap between Italian and English, thus increasing cross-lingual performance.


\paragraph{Sentence Reordering.}
Another typological feature which differs between languages, is the ordering of nouns and adjectives.
For instance, 
WALS shows that Romance languages such as French and Italians often use noun-adjective ordering, e.g., \textit{pomme rouge} in French, whereas the converse is used in English.
Additionally, languages differ in their ordering of subjects, objects, and verbs.
For instance, according to WALS, English firmly follows the subject-verb-object (SVO) structure, whereas there is no dominant order in German. 
We apply this reordering in order to decrease the linguistic gap between languages.
For instance, when considering English and French, we reverse all noun-adjective pairings from French to match English. 
This alignment is done while considering a dependency tree.
We re-align according to the typological features from WALS.
Since such feature annotations are available for a large amount of languages, and can be obtained automatically with high accuracy \citep{bjerva2019probabilistic}, we expect this method to scale to languages for which basic dependencies (such as noun-adjective attachment) can be obtained automatically. 
%
%
In \S\ref{sec:results}, we refer to the above re-alignment step as \metric{Text}.

\section{Experiments}\label{sec:experiments}

\subsection{Transfer tasks}
Cross-lingual embeddings are usually evaluated via zero-shot cross-lingual transfer for supervised text classification tasks, or via unsupervised cross-lingual textual similarity. 
For zero-shot transfer, fine-tuning of cross-lingual embeddings is done 
based on source language performance, and evaluation is performed on a held-out target language. 
This is, however, not likely to result in high quality target language embeddings and gives a false impression of cross-lingual abilities 
\citep{Libovick:2020}.
\citet{zhao:2020} 
use the more difficult task of reference-free machine translation evaluation (\rf{}) to expose 
limitations of cross-lingual encoders, i.e., 
a failure to properly represent fine-grained language aspects, which may be exploited by natural adversarial inputs such as word-by-word translations. 

\insertLangsDes

\paragraph{XNLI.} 
The goal of natural language inference (NLI) is to infer whether a premise sentence entails, contradicts, or is neutral towards a hypothesis sentence. 
\citet{conneau:2018} release a multilingual NLI corpus, where the English dev and test sets of the MultiNLI corpus \cite{williams:2018} are translated to 15 languages by 
crowd-workers.

\paragraph{\rf{}.}
This task  
evaluates the translation quality, i.e.~similarity of a target language translation and a source language sentence. 
Following \citet{zhao:2020}, we collect source language sentences with their system and reference translations, as well as human judgments from the WMT17 metrics shared task \citep{bojar:2017}, which contains predictions of 166 translation systems across 12 language pairs in WMT17.
Each language pair has approximately 3k source sentences, each associated with one human reference translation and with the automatic translations of participating systems. \wei{As in \citet{zhao:2019, zhao:2020}, we use the Earth Mover Distance 
to compute the 
distances between source sentence and target language translations, based on the semantic similarities of their contextualized cross-lingual embeddings. We refer to this score as XMoverScore \citep{zhao:2020} and report its Pearson
correlation with human judgments in our experiments.} 

\subsection{A Typologically Varied Language Sample}
We evaluate multilingual representations \wei{on two sets of languages: (1) a default language set with 4 languages from the official \xnli{} test sets and 2 languages from the WMT17 test sets; (2) a diagnostic language set which contains} 19 languages  with different levels of data resources from a typologically diverse sample\footnote{This sample was chosen as it yields a large typological variety, with representatives from several language families across the world.} covering 
five language families (each with at least three languages): Austronesian ($\alpha$), Germanic ($\beta$), Indo-Aryan ($\gamma$), Romance ($\delta$), and Uralic ($\eta$).
\wei{For \rf{}, we resort to pairs of translated source sentences and system translations. The former ones are translated from English human reference translations into 18 languages, obtained from Google Translate. For \xnli{}, we use translated test sets of all these languages from \cite{Hu:2020}}. 
Tab.~\ref{tab:langs} shows the overview of 19 languages which are labeled with 1) Similarity Level, i.e., the degree of similarity between target languages and English; and 2) Resource Level, i.e., the amount of data resources available in Wikipedia.





\subsection{Cross-lingual Encoders} 
Our goal is to improve the cross-lingual abilities of  
established contextualized cross-lingual embeddings.
These support around 100 languages and are 
pre-trained using 
monolingual 
language modeling.

\textbf{m-BERT} \citep{devlin:2019} is pre-trained on 
104 monolingual corpora from Wikipedia, with: 1) a vocabulary size of 110k; 2) language-specific tokenization tools for data pre-processing; and 3) two monolingual pre-training tasks: masked language modeling and next sentence prediction. 

\textbf{XLM-R} \citep{Conneau:2019a} is pre-trained on the CommonCrawl corpora of 100 languages, which contain more monolingual data than Wikipedia corpora, with 1) a vocabulary size of 250k; 2) a language-agnostic tokenization tool,
Sentence Piece \cite{kudo-richardson:2018} for data pre-processing; and 3) masked language modeling as the only monolingual pre-training task.
\wei{We apply \metric{norm}, \metric{text}, \metric{joint-align} and the combinations of these to the last layer of m-BERT and XLM-R, and report their performances on \xnli{} and \rf{} in \S\ref{sec:results}. 
To investigate the layer-wise effect of these modifications, we apply the modifications to individual layers and report the performances 
in \S\ref{sec:analysis}}. 
See the appendix for implementation details. 

\insertMainResults
\insertPerfHuman
\insertBatchNorm
\insertAblation
\insertPerfGainsByText
\insertLayerPlots
\insertLayerTwoLangs
\insertDistLanguages
\insertPerfGaps
\insertPerfGains

\section{Results}\label{sec:results}
Unlike re-mapping and vector space normalization, 
scaling input normalization to a large language sample is more difficult, as typological features differ across languages. 
Thus, we report the results of re-mapping and vector space normalization across 19 languages, while text normalization is evaluated on a smaller sample of languages.
\label{sec:results}




\paragraph{
Re-mapping and Vector Space Normalization.}
{In Tab.~\ref{tab:main_results}, we show results 
on machine translated test sets.}
The m-BERT space modified by \metric{Joint-Align $\oplus$ Norm} achieves consistent improvements 
on \rf{}  
(+10.1 points) and \xnli{} (+7.6 points) on average.  However, effects are different for XLM-R. The modified XLM-R outperforms 
the baseline XLM-R on \rf{} by the largest margin (+33.5 points), but the improvement is much smaller (+2.8 points) on \xnli{}. 
These gains are not an artefact of machine-translated test sets: we observe similar gains on 
human-translated data (see Fig.~\ref{fig:perf_human}).
 
In Tab. \ref{tab:ablation}, we tease apart the sources of improvements. Overall, the impacts of \metric{Norm} and \metric{Joint-Align} are substantial, and their effect is additive and sometimes even superadditive (e.g., m-BERT improves by 10.1 points on \rf{} when both \metric{Norm} and \metric{Joint-Align} are applied but only by 1.7 and 7.6 points individually). We note that the improvement from \metric{Norm} is 
more consistent across tasks and encoders, 
despite its simplicity and negligible cost. 
In contrast,  \metric{Joint-Align} 
has a positive effect for 
m-BERT but it does not help for XLM-R on the \xnli{} task, 
notwithstanding 
the minor difference of two encoders, e.g., much larger training data and a different tokenizer used in XLM-R. 
\sem{We believe the poor discriminative ability of XLM-R, \emph{viz.},\todo{SE: same as above} that it cannot distinguish word translations from random word pairs, leads to the inconsistent behavior of \metric{Joint-Align}. As a remedy, negative examples such as random pairs could be included in Eq.~(\ref{eqn:rotate}) during training so as to decrease the discriminative gap between m-BERT and XLM-R. 
This suggests that future research efforts should focus on the robustness of cross-lingual alignments.
\todo{SE: I think the last sentence is stylistically/grammatically odd and I would remove it}}

\paragraph{Batch vs. Layer Normalization.}

\sem{Unsurprisingly, the choice of batch size greatly influences \xnli{} performance when applying batch normalization for m-BERT and XLM-R (Fig.~\ref{fig:bn-ln}).\todo{SE: The font for XNLI was different elsewhere. It's missing here and everywhere else in the red colored part} We find that (i) the larger the batch size is, the smaller the impacts on \xnli{}, and (ii) a batch size of 8 performs best. Interestingly, layer normalization does not help for \xnli{}, even though it yields batch-independent statistics and is effective in stabilizing the training process~\citep{transformer:2017}. We note that per batch sequences with varying time steps (i.e., sentence length) are often padded with zero vectors in practice. This leads to inaccurate batch-independent statistics, as they are computed across all time steps, unlike batch normalization with per batch statistics for individual time steps. In addition to batch and layer normalizations, other normalizers \todo{SE: other what?} such as GroupNorm~\citep{groupnorm:2018} and PowerNorm~\citep{powernorm:2020} also receive attention in many communities.\todo{SE: "also involve a vivid discussion" is stylistically odd} This raises another concern towards a systematic investigation of normalizers for future work.
\todo{SE: the last sentence is stylistically odd}}

\paragraph{Linguistic Manipulation.} 
We apply input modifications to language pairs that contrast in either of three typological features: word contractions, noun-adjective and object-verb orderings. 
Fig.~\ref{fig:perf_gains_text} shows that reducing the linguistic gap between languages by \metric{TEXT} can sometimes lead to improvements (exemplified by m-BERT). 
Both French and Italian benefit considerably from both removing contractions (a) and reversing the order of adjectives and nouns (b), with no changes observed for Spanish.
As for reversing object-verb order (c), we again see improvements for 2 out of 3 languages. 
We hypothesize that the few cases without gains are due to the differing frequencies of occurrences of linguistic phenomena in XNLI and \rf{}. 
Another error source is the automatic analysis from \citet{straka:2016}, and improving this pre-processing step may further increase the performance of \metric{text}.

\section{Analysis}
\label{sec:analysis}


\vspace{1.3mm}
\noindent \textbf{(Q1)} \textit{How sensitive are normalization and post-hoc re-mapping across layers?}
\vspace{1.3mm}

\noindent\wei{In Fig.~\ref{fig:layers-de-en}, 
rather than checking results for the last layer only, we investigate improvements of our three modifications on \rf{} across all layers of and XLM-R
for one high-resource language pair (de-en) 
and one low-resource pair (jv-en) (see appendix). 
This reveals that, 
(1) for \xnli{}, applying \metric{joint-align}, \metric{norm} and \metric{text} to the last layer of m-BERT and XLM-R consistently results in the best performance. This indicates that the modifications to the last layer could be sufficient for \emph{supervised}  cross-lingual transfer tasks.
(2) 
However, 
the best results on \rf{} are oftentimes 
obtained from an intermediate layer.
Further, (3) we observe that \metric{joint-align} is not always effective, especially for XLM-R. E.g., it leads to the worst performance across all layers on \xnli{} for XLM-R, even below the baseline performance. (4) Reporting improvements on only the last layer may sometimes give a false and inflated impression, especially for \rf{}. 
E.g., the improvement (on \rf{}) of the three modifications 
over the original embeddings is almost 30 points for the last layer of XLMR, but it is less than 15 points for the penultimate layer.  
(5) Normalization and re-mapping typically stabilize layer-wise variances.}
(6) 
The gains of the three modifications are largely complementary across layers. (see also Fig.~\ref{fig:combination-across-layers}). 




\vspace{1.3mm}
\noindent \textbf{(Q2)} \textit{To what extent can our  modifications decrease the cross-lingual transfer gap, especially in low-resource scenarios and dissimilar languages?}
\vspace{1.3mm}

\noindent Tab.~\ref{tab:perf_gaps} shows that 
applying re-mapping and vector space normalization\footnote{
We do not apply text normalization in this setup because not all 
languages are covered in UDPipe.} 
to the last layer of m-BERT and XLM-R considerably reduces performance gaps 
\emph{viz.}: 
a) zero-shot transfer performance on \xnli{} 
between 
the English test set and the average performance on the other 18 languages; 
b) the difference between mono- and cross-lingual textual similarity on \rf{}, i.e., the difference between 
the average correlations of XMoverScore 
and human judgments on 19 languages 
obtained from \textit{reference-based}\footnote{
Reference-based evaluation assigns semantic similarity scores to pairs of system and reference translations in English.} and \textit{reference-free} MT evaluation setups.
Although smaller, the remaining gaps indicates further potential for improvement. 
Fig.~\ref{fig:perf_gains} shows the largest gains are on (1) low-resource languages and (2) languages 
most distant to English.


\vspace{1.3mm}
\noindent \textbf{(Q3)} 
\textit{Are our modifications to contextualized cross-lingual encoders language-agnostic?} 
\vspace{1.3mm}

\noindent Fig. \ref{fig:vis_lanaguges} (a) shows that 
the centroid vectors\footnote{\wei{Language centroids are representative (sentence) embeddings of languages averaged over monolingual Wikipedia data, as in \citet{Libovick:2019}. 
Although they use language families as a proxy, recent work shows that \textit{structural similarities} of languages are a more likely candidate \citep{bjerva:cl:2019}.
}} 
of languages within the same language family lie closely in the vector space, further showing that
language identity signals are stored in the m-BERT embeddings.
Fig. \ref{fig:vis_lanaguges} (b)+(c) shows that these signals are diminished in both re-aligned and normalized vector spaces, suggesting that the resulting embeddings in them are more language-agnostic.

\insertWALS

\vspace{1.3mm}
\noindent \textbf{(Q4)} \textit{To what extent do the typological relations learned from contextualized cross-lingual encoders deviate from those set out by expert typologists?}
\vspace{1.3mm}

\noindent Tab.~\ref{tab:wals} shows that language similarities, between English and other 18 languages, obtained from m-BERT and XLM-R have high correlations with structural language similarities\footnote{
The language similarity induced by WALS is the fraction of structural properties that have the same value in two languages among all 192 properties.} obtained from WALS\footnote{WALS covers approximately 200 linguistic features over 2500 languages, annotated by expert typologists.} via the syntactic features listed, indicating that \wei{language identifiers stored in the original embeddings are a good proxy for the annotated linguistic features.}
In contrast, this correlation is smaller in the modified embedding spaces, which we believe is because language identity is a much less prominent signal in them.

 



\section{Conclusion}
Cross-lingual systems 
show 
striking performance for transfer, but their success crucially relies on two constraints: the similarity between source and target languages and the size of pre-training corpora.
We 
comparatively evaluate three approaches to address these challenges, removing language-specific information from multilingual representations, thus learning language-agnostic representations. 
Our extensive experiments, based on a typologically broad sample of 19 languages, show that 
(vector space and input) normalization and re-mapping are oftentimes complementary approaches to improve cross-lingual performance, and that the popular approach of re-mapping leads to less consistent improvements than the much simpler and less costly normalization of vector representations. 
Input normalization yields benefits across a small sample of languages; 
further work is required for it to achieve consistent gains across a larger language sample.



\section*{Acknowledgments}
We thank the anonymous reviewers for their insightful comments and suggestions, which greatly improved the final version of the paper. 
This work has been supported by the German Research Foundation as part of the Research Training
Group Adaptive Preparation of Information from Heterogeneous Sources (AIPHES) at the Technische
Universit\"at Darmstadt under grant No. GRK 1994/1, as well as by the Swedish Research Council under grant agreement No 2019-04129.

\bibliography{anthology,acl2020}

\begin{thebibliography}{41}
\expandafter\ifx\csname natexlab\endcsname\relax\def\natexlab#1{#1}\fi

\bibitem[{Agi{\'c} and Vuli{\'c}(2019)}]{agic-vulic:2019}
{\v{Z}}eljko Agi{\'c} and Ivan Vuli{\'c}. 2019.
\newblock \href {https://doi.org/10.18653/v1/P19-1310} {{JW}300: A
  wide-coverage parallel corpus for low-resource languages}.
\newblock In \emph{Proceedings of the 57th Annual Meeting of the Association
  for Computational Linguistics}, pages 3204--3210, Florence, Italy.
  Association for Computational Linguistics.

\bibitem[{Aldarmaki and Diab(2019)}]{aldarmaki:2019}
Hanan Aldarmaki and Mona Diab. 2019.
\newblock \href {https://doi.org/10.18653/v1/N19-1391} {Context-aware
  cross-lingual mapping}.
\newblock In \emph{Proceedings of the 2019 Conference of the North {A}merican
  Chapter of the Association for Computational Linguistics: Human Language
  Technologies, Volume 1 (Long and Short Papers)}, pages 3906--3911,
  Minneapolis, Minnesota. Association for Computational Linguistics.

\bibitem[{Artetxe et~al.(2017)Artetxe, Labaka, and
  Agirre}]{artetxe-etal-2017-learning}
Mikel Artetxe, Gorka Labaka, and Eneko Agirre. 2017.
\newblock \href {https://doi.org/10.18653/v1/P17-1042} {Learning bilingual word
  embeddings with (almost) no bilingual data}.
\newblock In \emph{Proceedings of the 55th Annual Meeting of the Association
  for Computational Linguistics (Volume 1: Long Papers)}, pages 451--462,
  Vancouver, Canada. Association for Computational Linguistics.

\bibitem[{Ba et~al.(2016)Ba, Kiros, and Hinton}]{layernorm:2016}
Lei~Jimmy Ba, Jamie~Ryan Kiros, and Geoffrey~E. Hinton. 2016.
\newblock \href {http://arxiv.org/abs/1607.06450} {Layer normalization}.
\newblock \emph{CoRR}, abs/1607.06450.

\bibitem[{Bjerva and Augenstein(2018)}]{bjerva_augenstein:2018}
Johannes Bjerva and Isabelle Augenstein. 2018.
\newblock \href {https://doi.org/10.18653/v1/N18-1083} {From phonology to
  syntax: Unsupervised linguistic typology at different levels with language
  embeddings}.
\newblock In \emph{Proceedings of the 2018 Conference of the North {A}merican
  Chapter of the Association for Computational Linguistics: Human Language
  Technologies, Volume 1 (Long Papers)}, pages 907--916, New Orleans,
  Louisiana. Association for Computational Linguistics.

\bibitem[{Bjerva and Augenstein(2021)}]{bjerva-augenstein-2021-typological}
Johannes Bjerva and Isabelle Augenstein. 2021.
\newblock \href {https://www.aclweb.org/anthology/2021.eacl-main.38} {Does
  typological blinding impede cross-lingual sharing?}
\newblock In \emph{Proceedings of the 16th Conference of the European Chapter
  of the Association for Computational Linguistics: Main Volume}, pages
  480--486, Online. Association for Computational Linguistics.

\bibitem[{Bjerva et~al.(2019{\natexlab{a}})Bjerva, Kementchedjhieva, Cotterell,
  and Augenstein}]{bjerva2019probabilistic}
Johannes Bjerva, Yova Kementchedjhieva, Ryan Cotterell, and Isabelle
  Augenstein. 2019{\natexlab{a}}.
\newblock \href {https://doi.org/10.18653/v1/N19-1156} {A probabilistic
  generative model of linguistic typology}.
\newblock In \emph{Proceedings of the 2019 Conference of the North {A}merican
  Chapter of the Association for Computational Linguistics: Human Language
  Technologies, Volume 1 (Long and Short Papers)}, pages 1529--1540,
  Minneapolis, Minnesota. Association for Computational Linguistics.

\bibitem[{Bjerva et~al.(2019{\natexlab{b}})Bjerva, Östling, Veiga, Tiedemann,
  and Augenstein}]{bjerva:cl:2019}
Johannes Bjerva, Robert Östling, Maria~Han Veiga, Jörg Tiedemann, and
  Isabelle Augenstein. 2019{\natexlab{b}}.
\newblock \href {https://doi.org/10.1162/coli\_a\_00351} {{What Do Language
  Representations Really Represent?}}
\newblock \emph{Computational Linguistics}, 45(2):381--389.

\bibitem[{Bojar et~al.(2017)Bojar, Graham, and Kamran}]{bojar:2017}
Ond{\v{r}}ej Bojar, Yvette Graham, and Amir Kamran. 2017.
\newblock \href {https://doi.org/10.18653/v1/W17-4755} {Results of the {WMT}17
  metrics shared task}.
\newblock In \emph{Proceedings of the Second Conference on Machine
  Translation}, pages 489--513, Copenhagen, Denmark. Association for
  Computational Linguistics.

\bibitem[{Cao et~al.(2020)Cao, Kitaev, and Klein}]{Cao:2020}
Steven Cao, Nikita Kitaev, and Dan Klein. 2020.
\newblock \href {http://arxiv.org/abs/2002.03518} {Multilingual alignment of
  contextual word representations}.
\newblock \emph{CoRR}, abs/2002.03518.

\bibitem[{Conneau et~al.(2019)Conneau, Khandelwal, Goyal, Chaudhary, Wenzek,
  Guzm{\'{a}}n, Grave, Ott, Zettlemoyer, and Stoyanov}]{Conneau:2019a}
Alexis Conneau, Kartikay Khandelwal, Naman Goyal, Vishrav Chaudhary, Guillaume
  Wenzek, Francisco Guzm{\'{a}}n, Edouard Grave, Myle Ott, Luke Zettlemoyer,
  and Veselin Stoyanov. 2019.
\newblock \href {http://arxiv.org/abs/1911.02116} {Unsupervised cross-lingual
  representation learning at scale}.
\newblock \emph{CoRR}, abs/1911.02116.

\bibitem[{Conneau et~al.(2018)Conneau, Rinott, Lample, Williams, Bowman,
  Schwenk, and Stoyanov}]{conneau:2018}
Alexis Conneau, Ruty Rinott, Guillaume Lample, Adina Williams, Samuel Bowman,
  Holger Schwenk, and Veselin Stoyanov. 2018.
\newblock \href {https://doi.org/10.18653/v1/D18-1269} {{XNLI}: Evaluating
  cross-lingual sentence representations}.
\newblock In \emph{Proceedings of the 2018 Conference on Empirical Methods in
  Natural Language Processing}, pages 2475--2485, Brussels, Belgium.
  Association for Computational Linguistics.

\bibitem[{Devlin et~al.(2019)Devlin, Chang, Lee, and Toutanova}]{devlin:2019}
Jacob Devlin, Ming-Wei Chang, Kenton Lee, and Kristina Toutanova. 2019.
\newblock \href {https://doi.org/10.18653/v1/N19-1423} {{BERT}: Pre-training of
  deep bidirectional transformers for language understanding}.
\newblock In \emph{Proceedings of the 2019 Conference of the North {A}merican
  Chapter of the Association for Computational Linguistics: Human Language
  Technologies, Volume 1 (Long and Short Papers)}, pages 4171--4186,
  Minneapolis, Minnesota. Association for Computational Linguistics.

\bibitem[{Dryer and Haspelmath(2013)}]{wals}
Matthew~S. Dryer and Martin Haspelmath, editors. 2013.
\newblock \href {https://wals.info/} {\emph{WALS Online}}.
\newblock Max Planck Institute for Evolutionary Anthropology, Leipzig.

\bibitem[{Dyer et~al.(2013)Dyer, Chahuneau, and Smith}]{dyer:2013}
Chris Dyer, Victor Chahuneau, and Noah~A. Smith. 2013.
\newblock \href {https://www.aclweb.org/anthology/N13-1073} {A simple, fast,
  and effective reparameterization of {IBM} model 2}.
\newblock In \emph{Proceedings of the 2013 Conference of the North {A}merican
  Chapter of the Association for Computational Linguistics: Human Language
  Technologies}, pages 644--648, Atlanta, Georgia. Association for
  Computational Linguistics.

\bibitem[{Ethayarajh(2019)}]{ethayarajh:2019}
Kawin Ethayarajh. 2019.
\newblock \href {https://doi.org/10.18653/v1/D19-1006} {How contextual are
  contextualized word representations? comparing the geometry of {BERT},
  {ELM}o, and {GPT}-2 embeddings}.
\newblock In \emph{Proceedings of the 2019 Conference on Empirical Methods in
  Natural Language Processing and the 9th International Joint Conference on
  Natural Language Processing (EMNLP-IJCNLP)}, pages 55--65, Hong Kong, China.
  Association for Computational Linguistics.

\bibitem[{Hu et~al.(2020)Hu, Ruder, Siddhant, Neubig, Firat, and
  Johnson}]{Hu:2020}
Junjie Hu, Sebastian Ruder, Aditya Siddhant, Graham Neubig, Orhan Firat, and
  Melvin Johnson. 2020.
\newblock \href {http://arxiv.org/abs/2003.11080} {{XTREME:} {A} massively
  multilingual multi-task benchmark for evaluating cross-lingual
  generalization}.
\newblock \emph{CoRR}, abs/2003.11080.

\bibitem[{Koehn(2005)}]{koehn:2005}
Philipp Koehn. 2005.
\newblock Europarl: A parallel corpus for statistical machine translation.
\newblock In \emph{MT summit}, volume~5, pages 79--86. Citeseer.

\bibitem[{Kudo and Richardson(2018)}]{kudo-richardson:2018}
Taku Kudo and John Richardson. 2018.
\newblock \href {https://doi.org/10.18653/v1/D18-2012} {{S}entence{P}iece: A
  simple and language independent subword tokenizer and detokenizer for neural
  text processing}.
\newblock In \emph{Proceedings of the 2018 Conference on Empirical Methods in
  Natural Language Processing: System Demonstrations}, pages 66--71, Brussels,
  Belgium. Association for Computational Linguistics.

\bibitem[{Lample et~al.(2018)Lample, Conneau, Ranzato, Denoyer, and
  Jégou}]{lample2018word}
Guillaume Lample, Alexis Conneau, Marc'Aurelio Ranzato, Ludovic Denoyer, and
  Hervé Jégou. 2018.
\newblock Word translation without parallel data.
\newblock In \emph{ICLR}.

\bibitem[{Lauscher et~al.(2020)Lauscher, Ravishankar, Vulic, and
  Glavas}]{Lauscher:2020}
Anne Lauscher, Vinit Ravishankar, Ivan Vulic, and Goran Glavas. 2020.
\newblock \href {http://arxiv.org/abs/2005.00633} {From zero to hero: On the
  limitations of zero-shot cross-lingual transfer with multilingual
  transformers}.
\newblock \emph{CoRR}, abs/2005.00633.

\bibitem[{LeCun et~al.(1998)LeCun, Bottou, Orr, and M\"{u}ller}]{LeCun:1998}
Yann LeCun, L\'{e}on Bottou, Genevieve~B. Orr, and Klaus-Robert M\"{u}ller.
  1998.
\newblock Efficient backprop.
\newblock In \emph{Neural Networks: Tricks of the Trade, This Book is an
  Outgrowth of a 1996 NIPS Workshop}, page 9–50, Berlin, Heidelberg.
  Springer-Verlag.

\bibitem[{de~Lhoneux et~al.(2018)de~Lhoneux, Bjerva, Augenstein, and
  S{\o}gaard}]{delhoneux2018parameter}
Miryam de~Lhoneux, Johannes Bjerva, Isabelle Augenstein, and Anders S{\o}gaard.
  2018.
\newblock Parameter sharing between dependency parsers for related languages.
\newblock In \emph{Proceedings of the 2018 Conference on Empirical Methods in
  Natural Language Processing}, pages 4992--4997.

\bibitem[{Libovick{\'{y}} et~al.(2019)Libovick{\'{y}}, Rosa, and
  Fraser}]{Libovick:2019}
Jindrich Libovick{\'{y}}, Rudolf Rosa, and Alexander Fraser. 2019.
\newblock \href {http://arxiv.org/abs/1911.03310} {How language-neutral is
  multilingual bert?}
\newblock \emph{CoRR}, abs/1911.03310.

\bibitem[{Libovick{\'{y}} et~al.(2020)Libovick{\'{y}}, Rosa, and
  Fraser}]{Libovick:2020}
Jindrich Libovick{\'{y}}, Rudolf Rosa, and Alexander Fraser. 2020.
\newblock \href {http://arxiv.org/abs/2004.05160} {On the language neutrality
  of pre-trained multilingual representations}.
\newblock \emph{CoRR}, abs/2004.05160.

\bibitem[{Nivre et~al.(2020)Nivre, de~Marneffe, Ginter, Hajic, Manning,
  Pyysalo, Schuster, Tyers, and Zeman}]{Nivre:2020}
Joakim Nivre, Marie{-}Catherine de~Marneffe, Filip Ginter, Jan Hajic,
  Christopher~D. Manning, Sampo Pyysalo, Sebastian Schuster, Francis~M. Tyers,
  and Daniel Zeman. 2020.
\newblock \href {http://arxiv.org/abs/2004.10643} {Universal dependencies v2:
  An evergrowing multilingual treebank collection}.
\newblock \emph{CoRR}, abs/2004.10643.

\bibitem[{O'Horan et~al.(2016)O'Horan, Berzak, Vuli{\'c}, Reichart, and
  Korhonen}]{o2016survey}
Helen O'Horan, Yevgeni Berzak, Ivan Vuli{\'c}, Roi Reichart, and Anna Korhonen.
  2016.
\newblock Survey on the use of typological information in natural language
  processing.
\newblock \emph{arXiv preprint arXiv:1610.03349}.

\bibitem[{Pires et~al.(2019)Pires, Schlinger, and Garrette}]{pires:2019}
Telmo Pires, Eva Schlinger, and Dan Garrette. 2019.
\newblock \href {https://doi.org/10.18653/v1/P19-1493} {How multilingual is
  multilingual {BERT}?}
\newblock In \emph{Proceedings of the 57th Annual Meeting of the Association
  for Computational Linguistics}, pages 4996--5001, Florence, Italy.
  Association for Computational Linguistics.

\bibitem[{Ponti et~al.(2019)Ponti, O’horan, Berzak, Vuli{\'c}, Reichart,
  Poibeau, Shutova, and Korhonen}]{ponti2019modeling}
Edoardo~Maria Ponti, Helen O’horan, Yevgeni Berzak, Ivan Vuli{\'c}, Roi
  Reichart, Thierry Poibeau, Ekaterina Shutova, and Anna Korhonen. 2019.
\newblock Modeling language variation and universals: A survey on typological
  linguistics for natural language processing.
\newblock \emph{Computational Linguistics}, 45(3):559--601.

\bibitem[{R{\"u}ckl{\'e} et~al.(2018)R{\"u}ckl{\'e}, Eger, Peyrard, and
  Gurevych}]{rueckle:2018}
Andreas R{\"u}ckl{\'e}, Steffen Eger, Maxime Peyrard, and Iryna Gurevych. 2018.
\newblock \href {https://arxiv.org/abs/1803.01400} {Concatenated power mean
  word embeddings as universal cross-lingual sentence representations}.
\newblock \emph{arXiv}.

\bibitem[{Schuster et~al.(2019)Schuster, Ram, Barzilay, and
  Globerson}]{schuster:2019}
Tal Schuster, Ori Ram, Regina Barzilay, and Amir Globerson. 2019.
\newblock \href {https://doi.org/10.18653/v1/N19-1162} {Cross-lingual alignment
  of contextual word embeddings, with applications to zero-shot dependency
  parsing}.
\newblock In \emph{Proceedings of the 2019 Conference of the North {A}merican
  Chapter of the Association for Computational Linguistics: Human Language
  Technologies, Volume 1 (Long and Short Papers)}, pages 1599--1613,
  Minneapolis, Minnesota. Association for Computational Linguistics.

\bibitem[{Shen et~al.(2020)Shen, Yao, Gholami, Mahoney, and
  Keutzer}]{powernorm:2020}
Sheng Shen, Zhewei Yao, Amir Gholami, Michael Mahoney, and Kurt Keutzer. 2020.
\newblock \href {http://proceedings.mlr.press/v119/shen20e.html}
  {{P}ower{N}orm: Rethinking batch normalization in transformers}.
\newblock In \emph{Proceedings of the 37th International Conference on Machine
  Learning}, volume 119 of \emph{Proceedings of Machine Learning Research},
  pages 8741--8751. PMLR.

\bibitem[{Straka et~al.(2016)Straka, Haji{\v{c}}, and
  Strakov{\'a}}]{straka:2016}
Milan Straka, Jan Haji{\v{c}}, and Jana Strakov{\'a}. 2016.
\newblock {UDP}ipe: Trainable pipeline for processing {C}o{NLL}-u files
  performing tokenization, morphological analysis, {POS} tagging and parsing.
\newblock In \emph{Proceedings of the Tenth International Conference on
  Language Resources and Evaluation ({LREC}'16)}, Portoro{\v{z}}, Slovenia.

\bibitem[{Vaswani et~al.(2017)Vaswani, Shazeer, Parmar, Uszkoreit, Jones,
  Gomez, Kaiser, and Polosukhin}]{transformer:2017}
Ashish Vaswani, Noam Shazeer, Niki Parmar, Jakob Uszkoreit, Llion Jones,
  Aidan~N Gomez, \L~ukasz Kaiser, and Illia Polosukhin. 2017.
\newblock \href
  {https://proceedings.neurips.cc/paper/2017/file/3f5ee243547dee91fbd053c1c4a845aa-Paper.pdf}
  {Attention is all you need}.
\newblock In \emph{Advances in Neural Information Processing Systems},
  volume~30. Curran Associates, Inc.

\bibitem[{Wang and Eisner(2016)}]{galactic_dependencies}
Dingquan Wang and Jason Eisner. 2016.
\newblock \href {https://doi.org/10.1162/tacl_a_00113} {The galactic
  dependencies treebanks: Getting more data by synthesizing new languages}.
\newblock \emph{Transactions of the Association for Computational Linguistics},
  4:491--505.

\bibitem[{Williams et~al.(2018)Williams, Nangia, and Bowman}]{williams:2018}
Adina Williams, Nikita Nangia, and Samuel Bowman. 2018.
\newblock \href {https://doi.org/10.18653/v1/N18-1101} {A broad-coverage
  challenge corpus for sentence understanding through inference}.
\newblock In \emph{Proceedings of the 2018 Conference of the North {A}merican
  Chapter of the Association for Computational Linguistics: Human Language
  Technologies, Volume 1 (Long Papers)}, pages 1112--1122, New Orleans,
  Louisiana. Association for Computational Linguistics.

\bibitem[{Wu et~al.(2019)Wu, Conneau, Li, Zettlemoyer, and Stoyanov}]{Wu:2019}
Shijie Wu, Alexis Conneau, Haoran Li, Luke Zettlemoyer, and Veselin Stoyanov.
  2019.
\newblock \href {http://arxiv.org/abs/1911.01464} {Emerging cross-lingual
  structure in pretrained language models}.
\newblock \emph{CoRR}, abs/1911.01464.

\bibitem[{Wu and Dredze(2019)}]{wu:2019-beto}
Shijie Wu and Mark Dredze. 2019.
\newblock \href {https://doi.org/10.18653/v1/D19-1077} {Beto, bentz, becas: The
  surprising cross-lingual effectiveness of {BERT}}.
\newblock In \emph{Proceedings of the 2019 Conference on Empirical Methods in
  Natural Language Processing and the 9th International Joint Conference on
  Natural Language Processing (EMNLP-IJCNLP)}, pages 833--844, Hong Kong,
  China. Association for Computational Linguistics.

\bibitem[{Wu and He(2018)}]{groupnorm:2018}
Yuxin Wu and Kaiming He. 2018.
\newblock \href {https://doi.org/10.1007/978-3-030-01261-8\_1} {Group
  normalization}.
\newblock In \emph{Computer Vision - {ECCV} 2018 - 15th European Conference,
  Munich, Germany, September 8-14, 2018, Proceedings, Part {XIII}}, volume
  11217 of \emph{Lecture Notes in Computer Science}, pages 3--19. Springer.

\bibitem[{Zhao et~al.(2020)Zhao, Glava{\v{s}}, Peyrard, Gao, West, and
  Eger}]{zhao:2020}
Wei Zhao, Goran Glava{\v{s}}, Maxime Peyrard, Yang Gao, Robert West, and
  Steffen Eger. 2020.
\newblock \href {https://doi.org/10.18653/v1/2020.acl-main.151} {On the
  limitations of cross-lingual encoders as exposed by reference-free machine
  translation evaluation}.
\newblock In \emph{Proceedings of the 58th Annual Meeting of the Association
  for Computational Linguistics}, pages 1656--1671, Online. Association for
  Computational Linguistics.

\bibitem[{Zhao et~al.(2019)Zhao, Peyrard, Liu, Gao, Meyer, and
  Eger}]{zhao:2019}
Wei Zhao, Maxime Peyrard, Fei Liu, Yang Gao, Christian~M. Meyer, and Steffen
  Eger. 2019.
\newblock \href {https://doi.org/10.18653/v1/D19-1053} {{M}over{S}core: Text
  generation evaluating with contextualized embeddings and earth mover
  distance}.
\newblock In \emph{Proceedings of the 2019 Conference on Empirical Methods in
  Natural Language Processing and the 9th International Joint Conference on
  Natural Language Processing (EMNLP-IJCNLP)}, pages 563--578, Hong Kong,
  China. Association for Computational Linguistics.

\end{thebibliography}
\bibliographystyle{acl_natbib}

\clearpage

\appendix
\section{Appendix}

\subsection{Language Centroids}
We select 5k monolingual sentences from Wikipedia for 19 languages (each with at least 20 characters). Then, we normalize them by removing all punctuation, and use them to estimate language centroid vectors for each language. To do so, we first obtain their sentence embeddings by executing the mean pooling operation for the last layer of m-BERT (or XLM-R) contextualized word embeddings without [CLS] and [SEP] tokens involved. Then, we average these sentence embeddings to obtain language-specific centroid vectors.

\subsection{Our Modifications}
\paragraph{Re-mapping} We fine-tune m-BERT ($L= 12$, $H = 768$, $110M$ params) and XLM-R ($L= 12$, $H = 768$, $70M$ params) on the concatenated mutual word translations of 18 languages paired with English, using the loss function obtained as Eq.~\ref{eqn:rotate}. The mutual word translations are extracted with FastAlign \citep{dyer:2013} on parallel text from the combination of following publicly available parallel corpora.
\begin{itemize}
    \item Europarl \citep{koehn:2005}: We select 9 languages (German, Spanish, French, Italian, Dutch, Finnish, Hungarian, Portuguese, Estonian) out of 21 languages from Europarl. The size varies from 400k to 2M sentences depending on the language pair. We extract 100k parallel text for each language paired with English.
    \item JW300 \citep{agic-vulic:2019}: We select the remaining languages (Tagalog, Bengali, Javanese, Marathi, Hindi, Urdu, Afrikaans, Malay, Indonesian) out of 380 languages from JW300. The average size is 100K parallel sentences per language pair. We extract 100k parallel text based on sampling for each language paired with English. 
\end{itemize}


\paragraph{Input Normalization} 
In the \rf{} setup, we do not modify system translations (in English), and instead manipulate source language texts. For XNLI, we manipulate both premise and hypothesis texts. To examine the impact of linguistic changes to cross-lingual transfer, we remove all punctuation from input texts. We extract word and lemma forms, universal part-of-speech (POS) tags, morphological features and universal dependency relations from input texts using UDPipe, which is a pipeline trained on UD treebank 2.5. Each orthographic token is split into several tokens that can be directly obtained from the corresponding word forms. To reverse noun-adjective and object-verb ordering, we use a simple rule-based strategy based on universal POS tags and universal dependency relations.

\insertLayerEnJv

\end{document}